\documentclass[runningheads]{llncs}
\usepackage{graphicx}
\usepackage{comment}
\usepackage{amsmath,amssymb} 
\usepackage{color}

\usepackage{multirow}
\usepackage{lipsum,graphicx}
\usepackage{capt-of}
\usepackage{booktabs}
\usepackage{varwidth}
\usepackage{fancyhdr,graphicx,amsmath,amssymb}
\usepackage[ruled,vlined]{algorithm2e}
\usepackage{caption}
\usepackage{subcaption}
\usepackage{subcaption}
\usepackage{graphicx}
\usepackage{lipsum}
\usepackage{graphicx}
\usepackage{dblfloatfix}    
\usepackage{stackengine}
\usepackage{amsmath}
\usepackage{bm}
\usepackage{color}

\DeclareMathOperator*{\argmax}{arg\,max}

\begin{document}
\pagestyle{headings}
\mainmatter
\def\ECCVSubNumber{2826}  

\title{Associative Alignment \\ for Few-shot Image Classification} 

\titlerunning{ECCV-20 submission ID \ECCVSubNumber} 
\authorrunning{ECCV-20 submission ID \ECCVSubNumber} 
\author{Anonymous ECCV submission}
\institute{Paper ID \ECCVSubNumber}

\titlerunning{Associative Alignment for Few-shot Image Classification}
%

\author{Arman Afrasiyabi$^*$, 
Jean-Fran\c{c}ois Lalonde$^*$, 
Christian Gagn\'e$^{* \dag}$ 
}
\authorrunning{A. Afrasiyabi et al.}
%
\institute{$^*$Universit\'e Laval, 
$^\dag$Canada CIFAR AI Chair, Mila   \\
\email{arman.afrasiyabi.1@ulaval.ca}\\
\email{\{jflalonde,christian.gagne\}@gel.ulaval.ca}  
\texttt{\url{https://lvsn.github.io/associative-alignment/}}}
\maketitle

\begin{abstract}

Few-shot image classification aims at training a model from only a few examples for each of the ``novel'' classes. This paper proposes the idea of associative alignment for leveraging part of the base data by aligning the novel training instances to the closely related ones in the base training set. This expands the size of the effective novel training set by adding extra ``related base'' instances to the few novel ones, thereby allowing a constructive fine-tuning. We propose two associative alignment strategies: 1) a metric-learning loss for minimizing the distance between related base samples and the centroid of novel instances in the feature space, and 2) a conditional adversarial alignment loss based on the Wasserstein distance. Experiments on four standard datasets and three backbones demonstrate that combining our centroid-based alignment loss results in absolute accuracy improvements of 4.4\%, 1.2\%, and 6.2\% in 5-shot learning over the state of the art for object recognition, fine-grained classification, and cross-domain adaptation, respectively.

\end{abstract}

\section{Introduction}
Despite recent progress, generalizing on new concepts with little supervision is still a challenge in computer vision. 
In the context of image classification, few-shot learning aims to obtain a model that can learn to recognize novel image classes when very few training examples are available.

Meta-learning~\cite{finn2017model,ravi2016optimization,snell2017prototypical,vinyals2016matching} is a possible approach to achieve this, by extracting common knowledge from a large amount of labeled data 
(the ``base'' classes) to train a model that can then learn to classify images from ``novel'' concepts with only a few examples. This is achieved by repeatedly sampling small subsets from the large pool of base images, effectively simulating the few-shot scenario.   
\begin{figure}
    \centering
    \footnotesize
    \setlength{\tabcolsep}{1pt}
    \begin{tabular}{cc}
    \includegraphics[height=4.0cm]{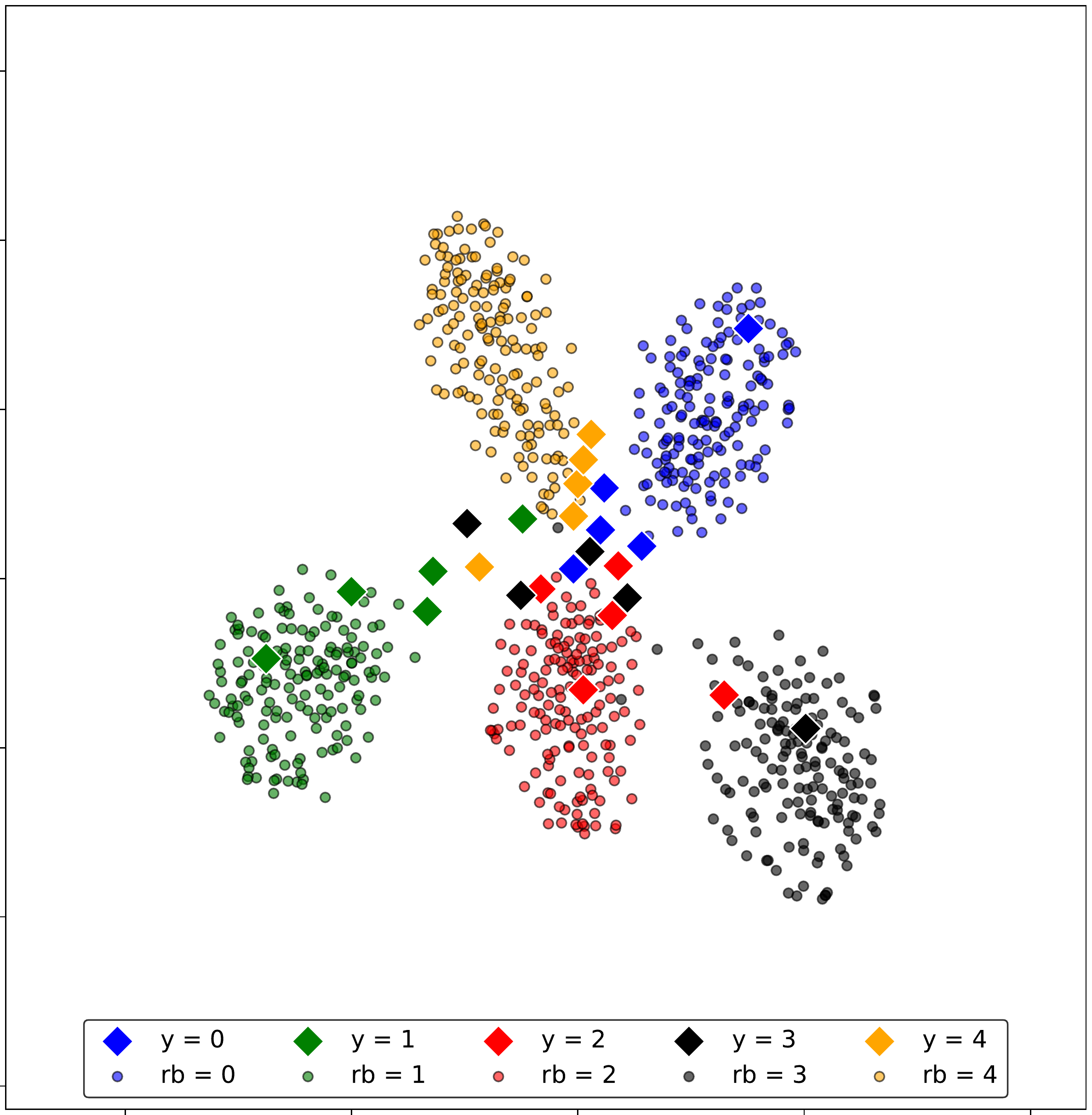} & 
    \includegraphics[height=4.0cm]{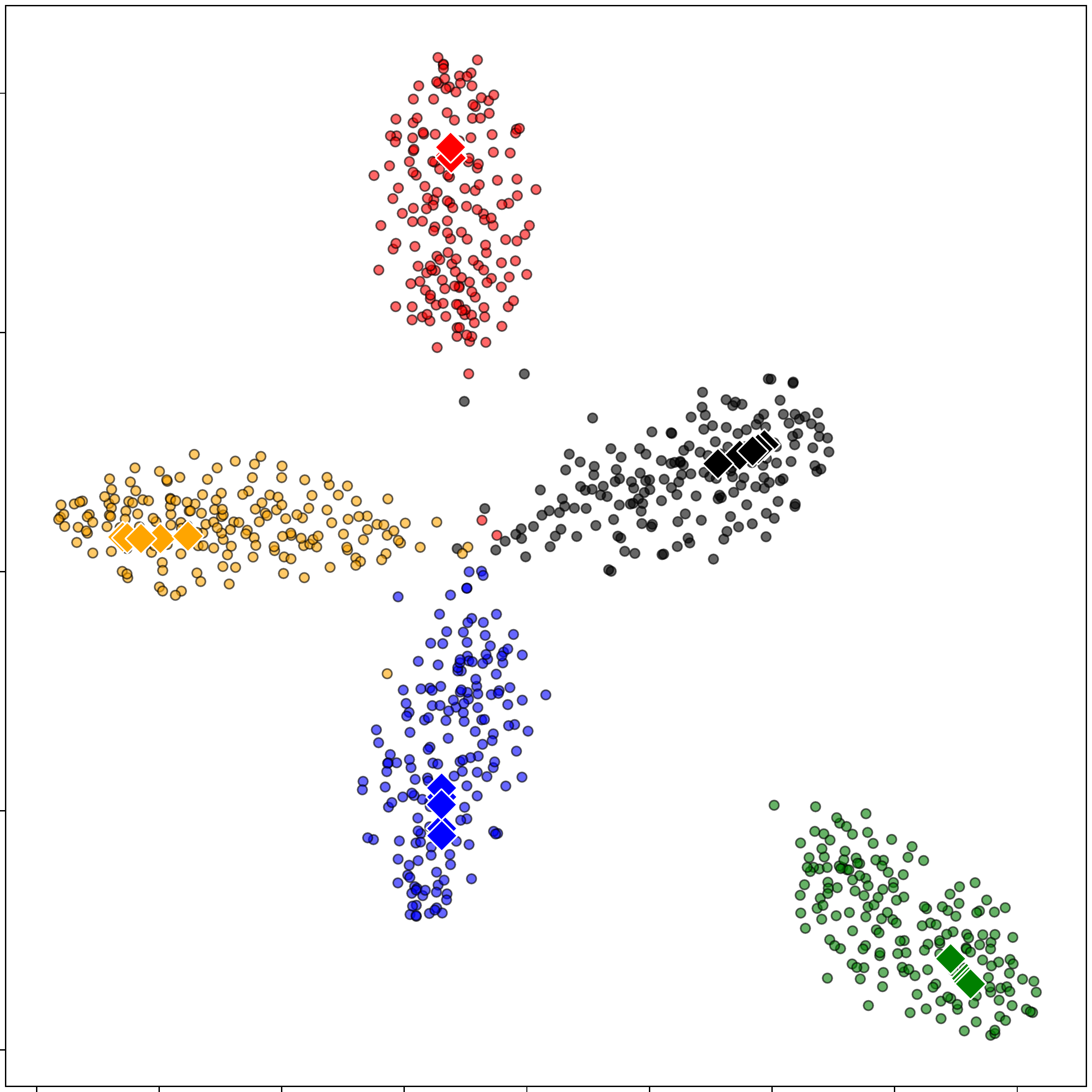} \\
    (a) before alignment & 
    (b) after alignment
    \end{tabular}
    \caption{The use of many related bases (circles) in addition to few novel classes samples (diamonds) allows better discriminative models: (a) using directly related bases may not properly capture the novel classes; while (b) aligning both related base and novel training instances (in the feature space) provides more relevant training data for classification. Plots are generated with t-SNE~\cite{maaten2008visualizing} applied to the ResNet-18 feature embedding before (a) and after (b) the application of the centroid alignment. Points are color-coded by class. 
    }
    \label{fig:idea}
\end{figure}
Standard transfer learning has also been explored as an alternative method~\cite{chen2019closer,gidaris2018dynamic,Qi_2018_CVPR}. The idea is to pre-train a network on the base samples and then fine-tune the classification layer on the novel examples. Interestingly, Chen \emph{et al.}~\cite{chen2019closer} demonstrated that doing so performs on par with more sophisticated meta-learning strategies. It is, however, necessary to freeze the feature encoder part of the network when fine-tuning on the novel classes since the network otherwise overfits the novel examples. We hypothesize that this hinders performance and that gains could be made if the entire network is adapted to the novel categories.

In this paper, we propose an approach that simultaneously prevents overfitting without restricting the learning capabilities of the network for few-shot image classification. Our approach relies on the standard transfer learning strategy~\cite{chen2019closer} as a starting point, but subsequently exploits base categories that are most similar (in the feature space) to the few novel samples to effectively provide additional training examples. We dub these similar categories the ``related base'' classes. Of course, the related base classes represent different concepts than the novel classes, so fine-tuning directly on them could confuse the network (see fig.~\ref{fig:idea}-(a)). The key idea of this paper is to \emph{align}, in feature space, the novel examples with the related base samples (fig.~\ref{fig:idea}-(b)). 

To this end, we present two possible solutions for associative alignment: by 1) centroid alignment, inspired by ProtoNet~\cite{snell2017prototypical}, benefits from explicitly shrinking the intra-class variations and is more stable to train, but makes
the assumption that the class distribution is well-approximated by a single mode. Adversarial alignment, inspired by WGAN~\cite{arjovsky2017wasserstein}, does not make that assumption, but its train complexity is greater
due to the critic network.
We demonstrate, through extensive experiments, that our centroid-based alignment procedure achieves state-of-the-art performance in few-shot classification on several standard benchmarks. Similar results are obtained by our adversarial alignment, which shows the effectiveness of our associative alignment approach. 

We present the following contributions. 
First, we propose two approaches for aligning novel to related base classes in the feature space, allowing for effective training of entire networks for few-shot image classification. Second, we introduce a strong baseline that combines standard transfer learning~\cite{chen2019closer} with an additive angular margin loss~\cite{deng2018arcface}, along with early stopping to regularize the network while pre-training on the base categories. We find that this simple baseline actually improves on the state of the art, in the best case by 3\% in overall accuracy. Third, we demonstrate through extensive experiments---on four standard datasets and using three well-known backbone feature extractors---that our proposed centroid alignment significantly outperforms the state of the art in three types of scenarios: 
generic object recognition (gain of 1.7\%, 4.4\% 2.1\% in overall accuracy for 5-shot on \textit{mini}-ImageNet, tieredImageNet and FC100 respectively), fine-grained classification (1.2\% on CUB), and cross-domain adaptation (6.2\% from \textit{mini}-ImageNet to CUB) using the ResNet-18 backbone. 

\section{Related work}


The main few-shot learning approaches can be broadly categorized into meta-learning and standard transfer learning. In addition, data augmentation and regularization techniques (typically in meta-learning) have also been used for few-shot learning. We briefly review relevant works in each category below. Note that several different computer vision problems such as object counting~\cite{Zhao_2018_ECCV}, video classification~\cite{Zhu_2018_ECCV}, motion prediction~\cite{Gui_2018_ECCV}, and object detection~\cite{wang2019meta} have been framed as few-shot learning. Here, we mainly focus on works from the image
classification literature.

\paragraph{Meta-learning} This family of approaches frames few-shot learning in the form of episodic training~\cite{dhillon2019baseline,finn2017model,ravi2016optimization,rusu2018meta,snell2017prototypical,vilalta2002perspective,wang2019meta,yoon2019tapnet}. An episode is defined by pretending to be in a few-shot regime while training on the base categories, which are available in large quantities. Initialization- and metric-based approaches are two variations on the episodic training scheme relevant for this work. 
Initialization-based methods~\cite{finn2017model,finn2018probabilistic,kim2018bayesian} learn an initial model 
able to adapt to few novel samples with a small number of gradient steps. In contrast, our approach performs a larger number of updates, but requires that the alignment be maintained between the novel samples and their related base examples. 
Metric-based approaches~\cite{bertinetto2018metalearning,garcia2017few,kim2019variational,Li_2019_CVPR,lifchitz2019dense,oreshkin2018tadam,snell2017prototypical,sung2018learning,tseng2020cross,vinyals2016matching,wertheimer2019few,zhang2019variational} learn a metric with the intent of reducing the intra-class variations while training on base categories. For example, ProtoNet~\cite{snell2017prototypical} were proposed to learn a feature space where instances of a given class are located close to the corresponding prototype (centroid), allowing accurate distance-based classification. 
Our centroid alignment strategy borrows from such distance-based criteria but uses it to match the distributions in the feature space 
instead of building a classifier. 

\paragraph{Standard transfer learning} The strategy behind this method is to pre-train a network on the base classes and subsequently fine-tune it on the novel examples~\cite{chen2019closer,gidaris2018dynamic,Qi_2018_CVPR}. Despite its simplicity, Chen \emph{et al.}~\cite{chen2019closer} recently demonstrated that such an approach could result in similar generalization performance compared to meta-learning when deep backbones are employed as feature extractors. However, they have also shown that the weights of the pre-trained feature extractor must remain frozen while fine-tuning due to the propensity for overfitting. Although the training procedure we are proposing is similar to standard fine-tuning in base categories, our approach allows the training of the entire network, thereby increasing the learned model capacity while improving classification accuracy. 

\paragraph{Regularization trick} 
Wang \emph{et al.}~\cite{wang2016learning-eccv} proposed regression networks for regularization purposes by refining the parameters of the fine-tuning model to be close to the pre-trained model.
More recently, Lee \emph{et al.}~\cite{lee2019meta} exploited the implicit differentiation of a linear classifier with hinge loss and $\mathcal{L}_2$ regularization to the CNN-based feature learner. Dvornik \emph{et al.}~\cite{dvornik2019diversity} uses an ensemble of networks to decrease the classifiers variance.

\paragraph{Data augmentation} Another family of techniques relies on additional data for training in a few-shot regime, most of the time following a meta-learning training procedure~\cite{Chen_2019_CVPR,Chu_2019_CVPR,gao2018low,gidaris2019generating,hariharan2017low,mehrotra2017generative,schwartz2018delta,wang2018low,Zhang_2019_CVPR,zhang2017mixup}. Several ways of doing so have been proposed, including Feature Hallucination (FH)~\cite{hariharan2017low}, which learns mappings between examples with an auxiliary generator that then hallucinates extra training examples (in the feature space). Subsequently, Wang \emph{et al.}~\cite{wang2018low} proposed to use a GAN for the same purpose, and thus address the poor generalization of the FH framework. Unfortunately, it has been shown that this approach suffers from mode collapse~\cite{gao2018low}. Instead of generating artificial data for augmentation, others  have proposed methods to take advantage of additional unlabeled data \cite{gidaris2019boosting,ren2018meta,li2019learning,wang2016learning}. Liu \emph{et al.}~\cite{Liu_2019_ICCV} propose to propagate labels from few labeled data to many unlabeled data, akin to our detection of related bases. We also rely on more data for training, but in contrast to these approaches, our method does not need any new data, nor does it require to generate any. Instead, we exploit the data that is \emph{already available} in the base domain and align the novel domain to the relevant base samples through fine-tuning. 
 
Previous work has also exploited base training data, most related to ours are the works of  \cite{Chen_2019_CVPR} and \cite{Lim2011Transfer}. 
Chen \emph{et al.}~\cite{Chen_2019_CVPR} propose to use an embedding and deformation sub-networks to leverage additional training samples, whereas we rely on a single feature extractor network which is much simpler to implement and train. Unlike random base example sampling~\cite{Chen_2019_CVPR} for interpolating novel example deformations in the image space, we propose to borrow the internal distribution structure of the detected related classes in feature space. Besides, our alignment strategies introduce extra criteria to keep the focus of the learner on the novel classes, which prevents the novel classes from becoming outliers.
Focused on object detection, Lim \emph{et al.}~\cite{Lim2011Transfer} proposes a model to search similar object categories using a sparse grouped Lasso framework.  Unlike~\cite{Lim2011Transfer}, we propose and evaluate two associative alignments in the context of few-shot image classification. 

From the alignment perspective, our work is related to Jiang \emph{et al.}~\cite{Jiang_2018_ECCV} which stays in the context of zero-shot learning, and proposes a coupled dictionary matching in visual-semantic structures to find matching concepts. In contrast, we propose associative base-novel class alignments along with two strategies for enforcing the unification of the related concepts.

\section{Preliminaries}
\label{sec:preliminaries}

Let us assume that we have a large base dataset $\mathcal{X}^b=\{(\mathbf{x}_i^b,y_i^b)\}_{i=1}^{N^b}$, where $\mathbf{x}_i^b\in\mathbb{R}^d$ is the $i$-th data instance of the set and $y_i^b\in\mathcal{Y}^b$ is the corresponding class label. We are also given a small amount of novel class data $\mathcal{X}^n=\{(\mathbf{x}_i^n,y_i^n)\}_{i=1}^{N^n}$, with labels $y_i^n\in\mathcal{Y}^n$ from a set of distinct classes $\mathcal{Y}^n$. Few-shot classification aims to train a classifier with only a few examples from each of the novel classes (e.g., 5 or even just 1). In this work, we used the standard transfer learning strategy of Chen \emph{et al.}~\cite{chen2019closer}, which is organized into the following two stages. 

\paragraph{Pre-training stage} The learning model is a neural network composed of a feature extractor $f(\cdot|\theta)$, parameterized by $\theta$, followed by a linear classifier $c(\mathbf{x}|\mathbf{W})\equiv\mathbf{W}^\top f(\mathbf{x}|\theta)$, described by matrix $\mathbf{W}$, ending with a scoring function such as softmax to produce the output. The network is trained from scratch on examples from the base categories $\mathcal{X}^b$.

\paragraph{Fine-tuning stage} In order to adapt the network to the novel classes, the network is subsequently fine-tuned on the few examples from $\mathcal{X}^n$. Since overfitting is likely to occur if all the network weights are updated, the feature extractor weights $\theta$ are frozen, with only the classifier weights $\mathbf{W}$ being updated in this stage. 

\section{Associative alignment}
\label{sec:alignment}
%
Freezing the feature extractor weights $\theta$ indeed reduces overfitting, but also limits the learning capacity of the model. In this paper, we strive for the best of both worlds and present an approach which controls overfitting while maintaining the original learning capacity of the model. We borrow the internal distribution structure of a subset of \emph{related} base categories, $\mathcal{X}^{rb}\subset\mathcal{X}^b$. 
To account for the discrepancy between the novel and related base classes, we propose to \emph{align} the novel categories to the related base categories in feature space. Such a mapping allows for a bigger pool of training data while making instances of these two sets more coherent. Note that, as opposed to~\cite{Chen_2019_CVPR}, we do not modify the related base instances in any way: we simply wish to align novel examples to the distributions of their related class instances. 
 
In this section, we first describe how the related base classes are determined. Then, we present our main contribution: the ``centroid associative alignment'' method, which exploits the related base instances to improve classification performance on novel classes. We conclude by presenting an alternative associative alignment strategy, which relies on an adversarial framework. 

\subsection{Detecting the related bases}
\begin{figure}[!t]
    \centering
    \footnotesize
    \setlength{\tabcolsep}{1pt}
    \begin{tabular}{cccccc} 
    \rotatebox{90}{nov.} & 
    \includegraphics[width=.19\linewidth]{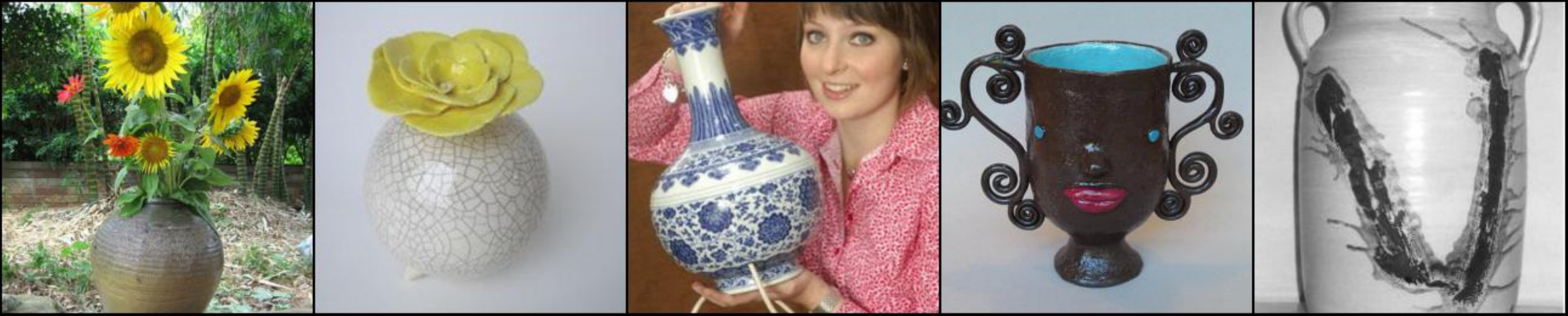} &  
    \includegraphics[width=.19\linewidth]{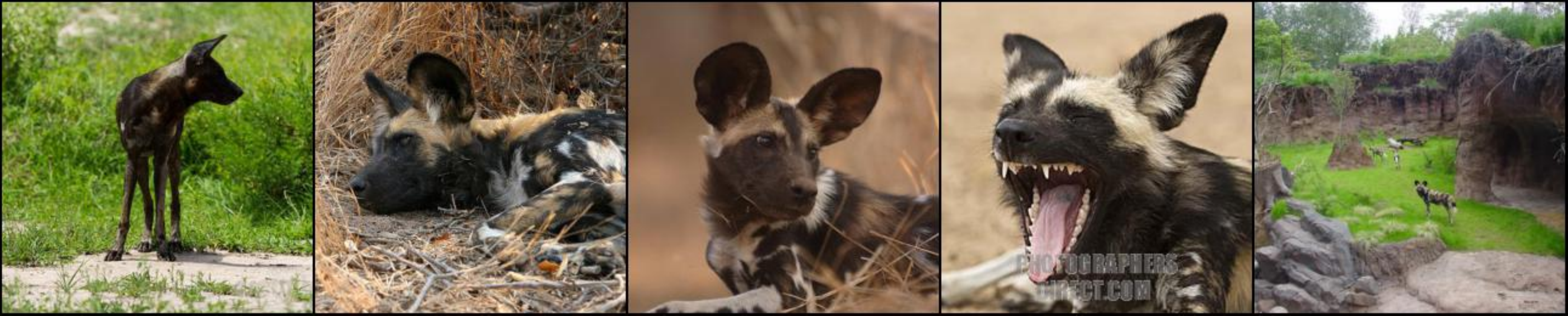} &  
    \includegraphics[width=.19\linewidth]{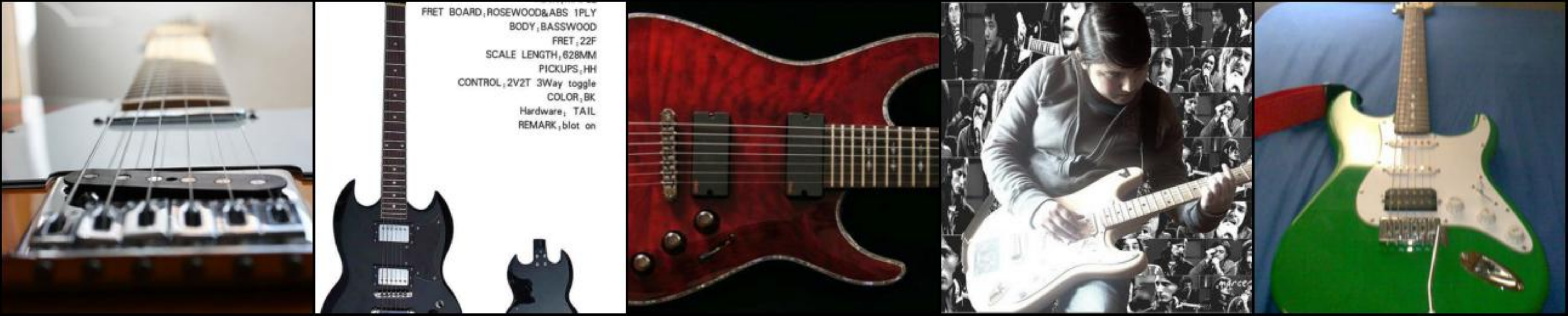} & 
    \includegraphics[width=.19\linewidth]{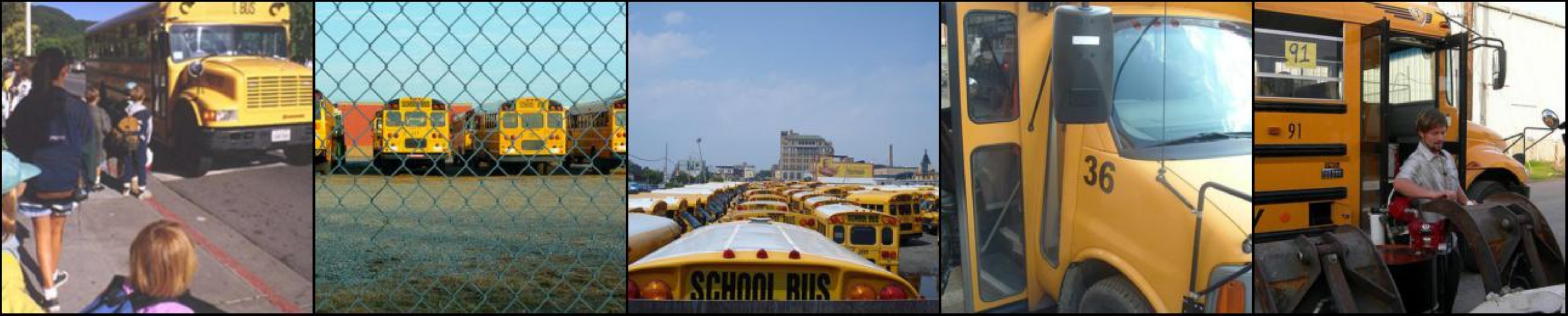} &
    \includegraphics[width=.19\linewidth]{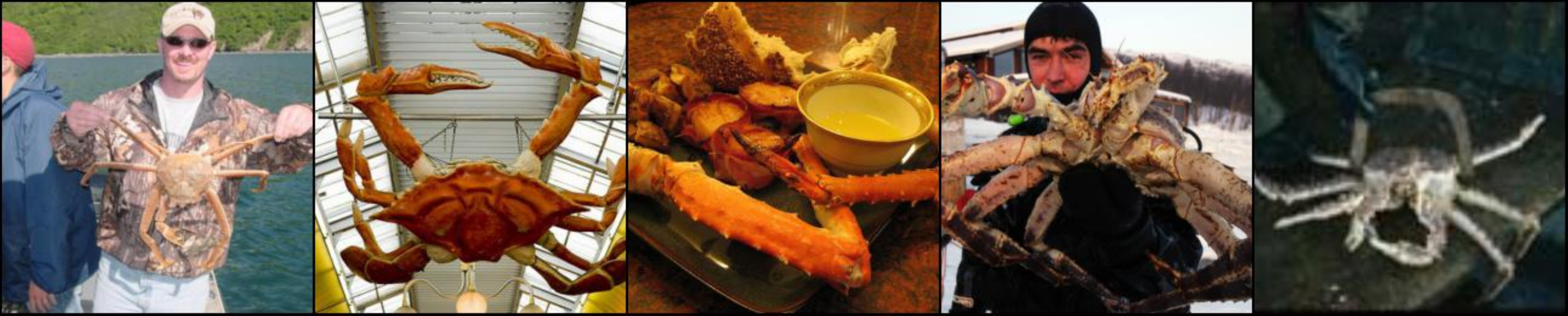} \\
    \rotatebox{90}{rel. bas.} & 
    \includegraphics[width=.19\linewidth]{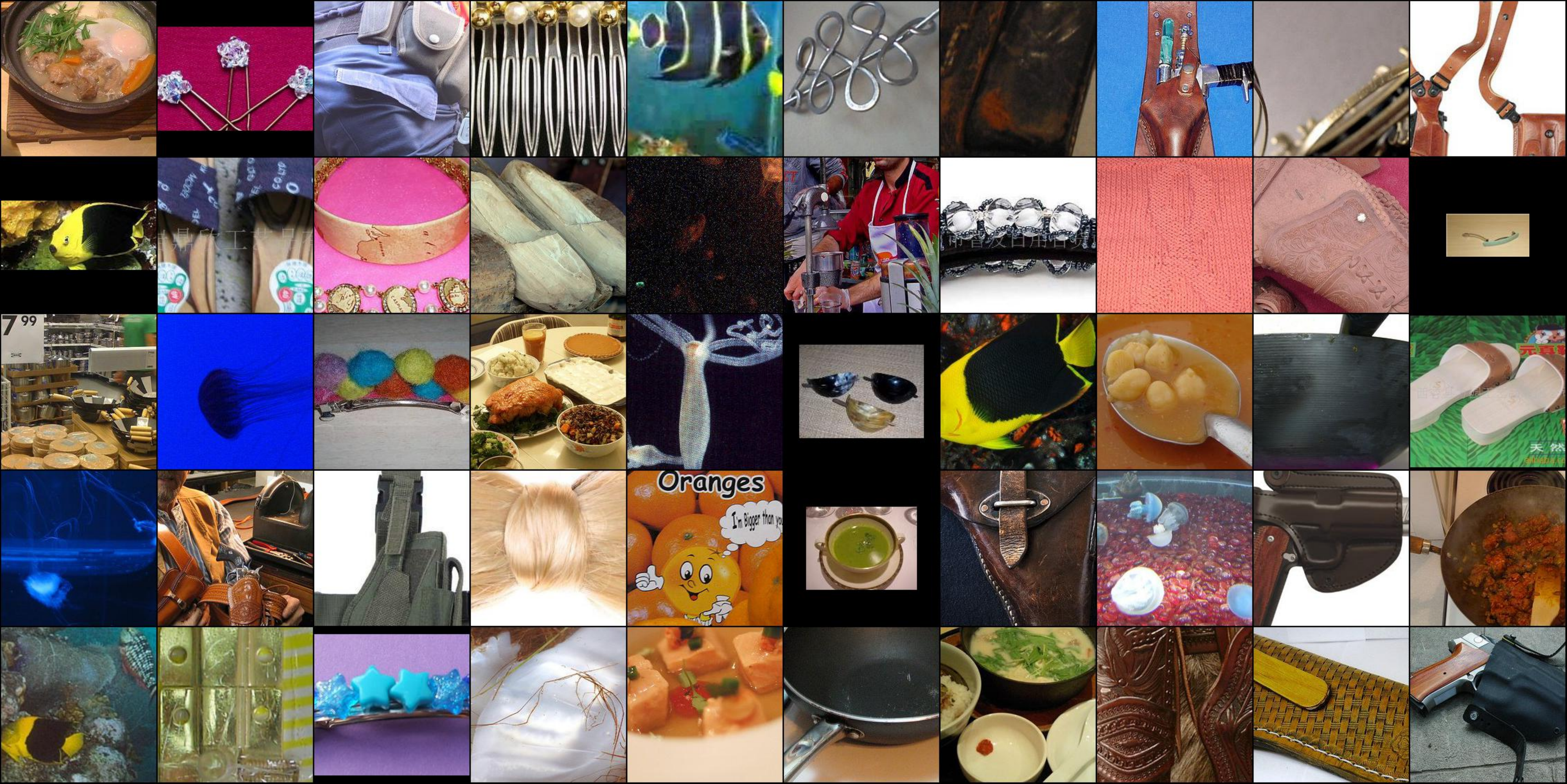} &
    \includegraphics[width=.19\linewidth]{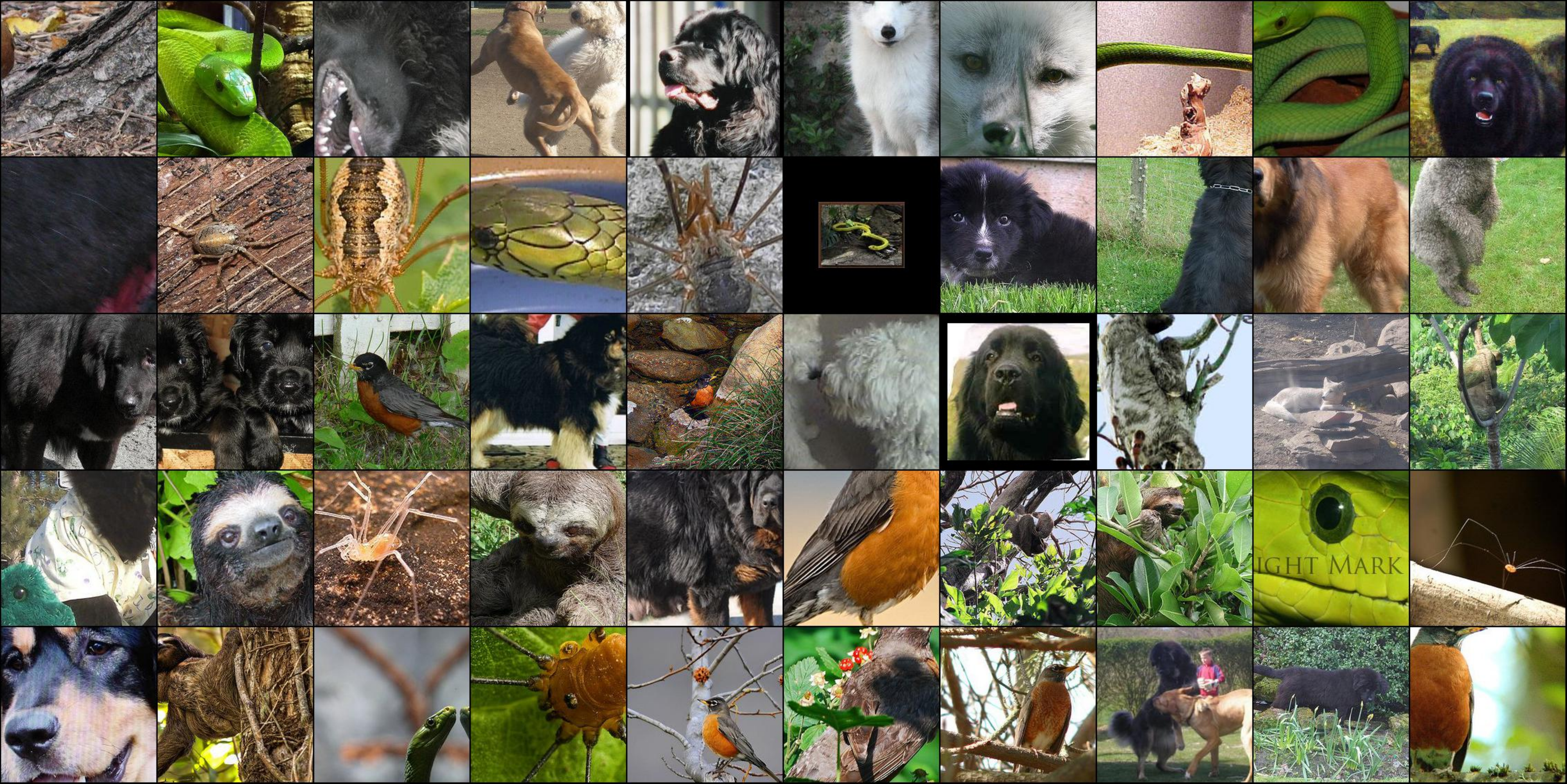} &
    \includegraphics[width=.19\linewidth]{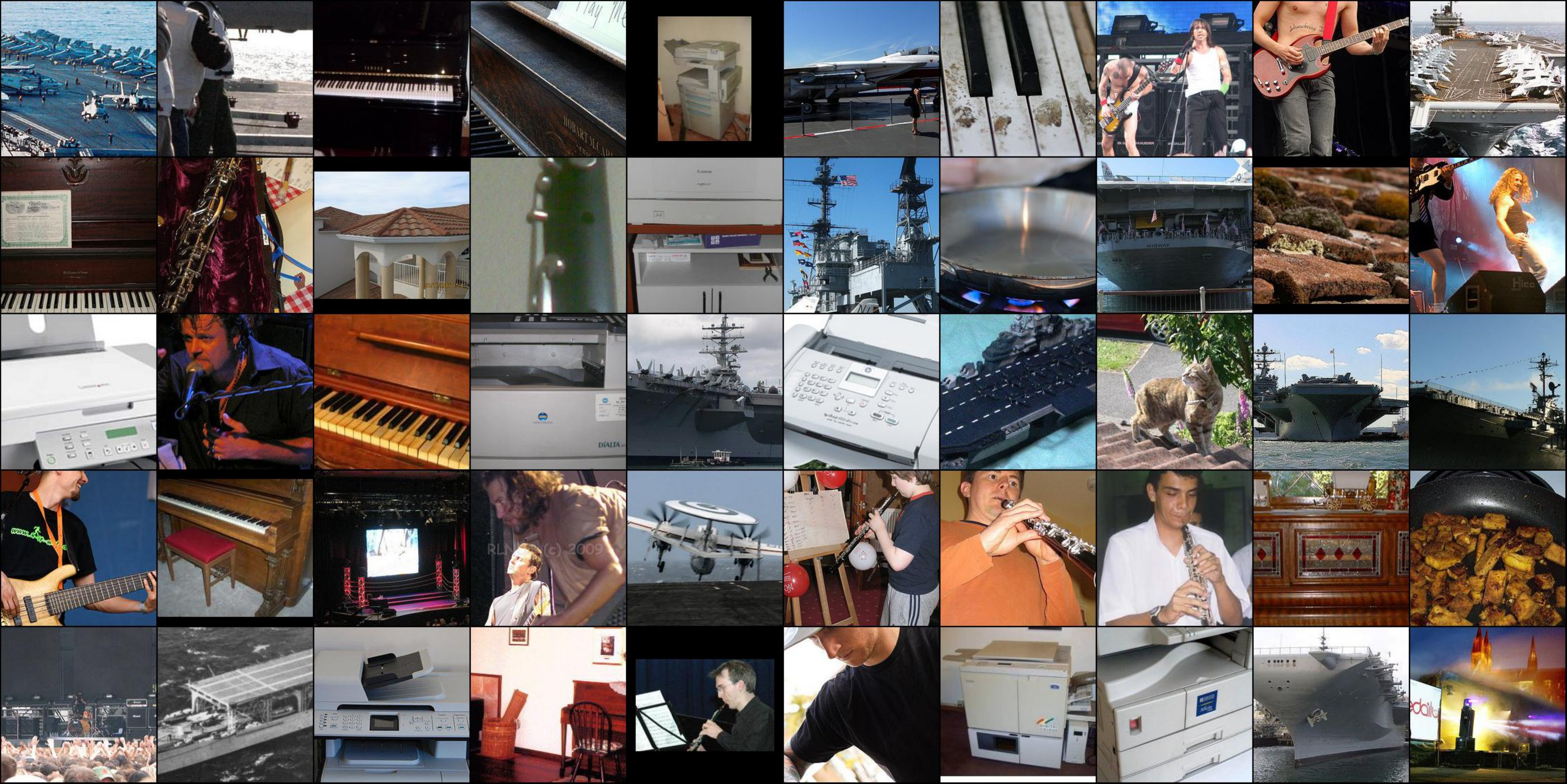} &
    \includegraphics[width=.19\linewidth]{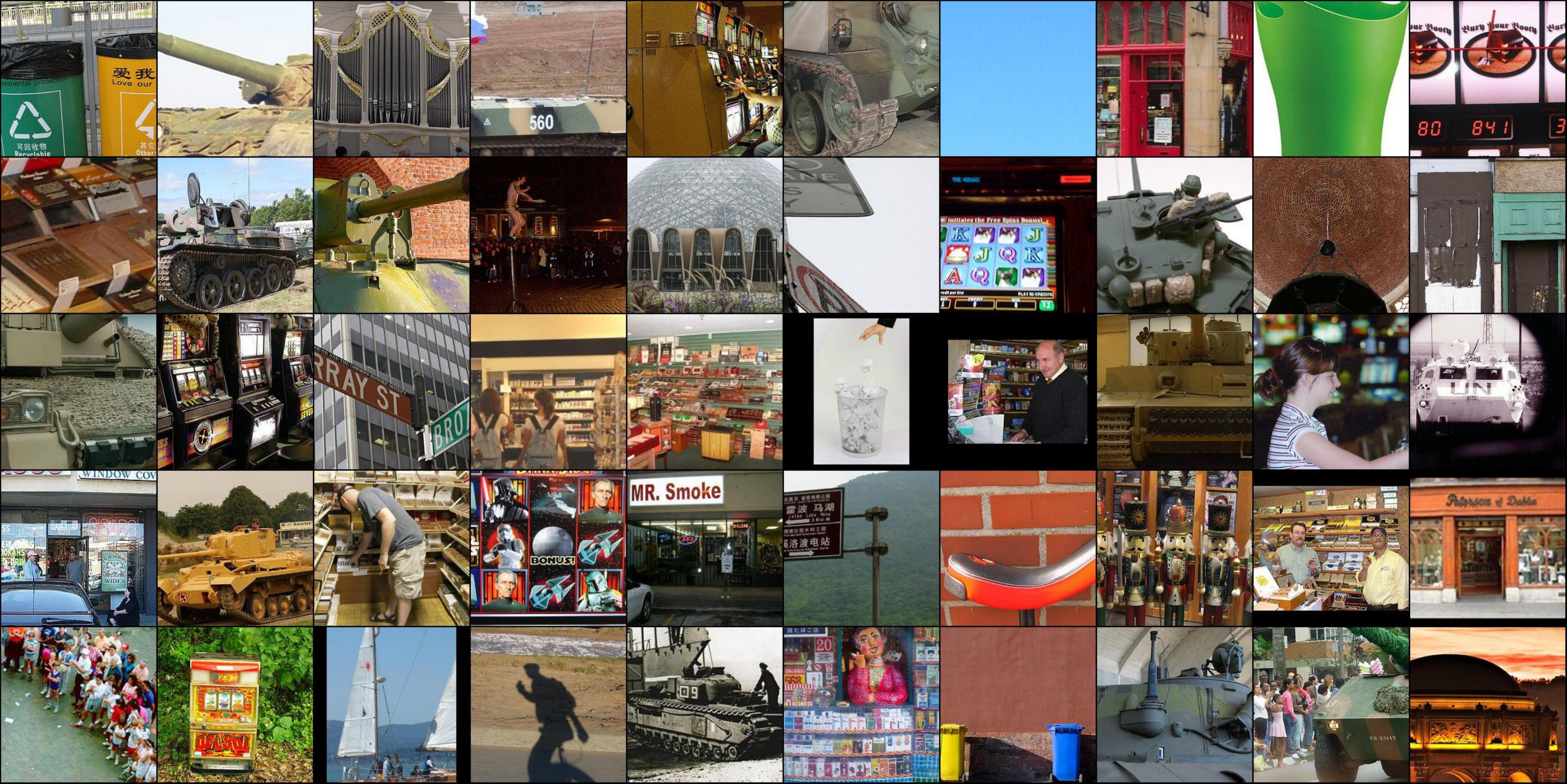} &
    \includegraphics[width=.19\linewidth]{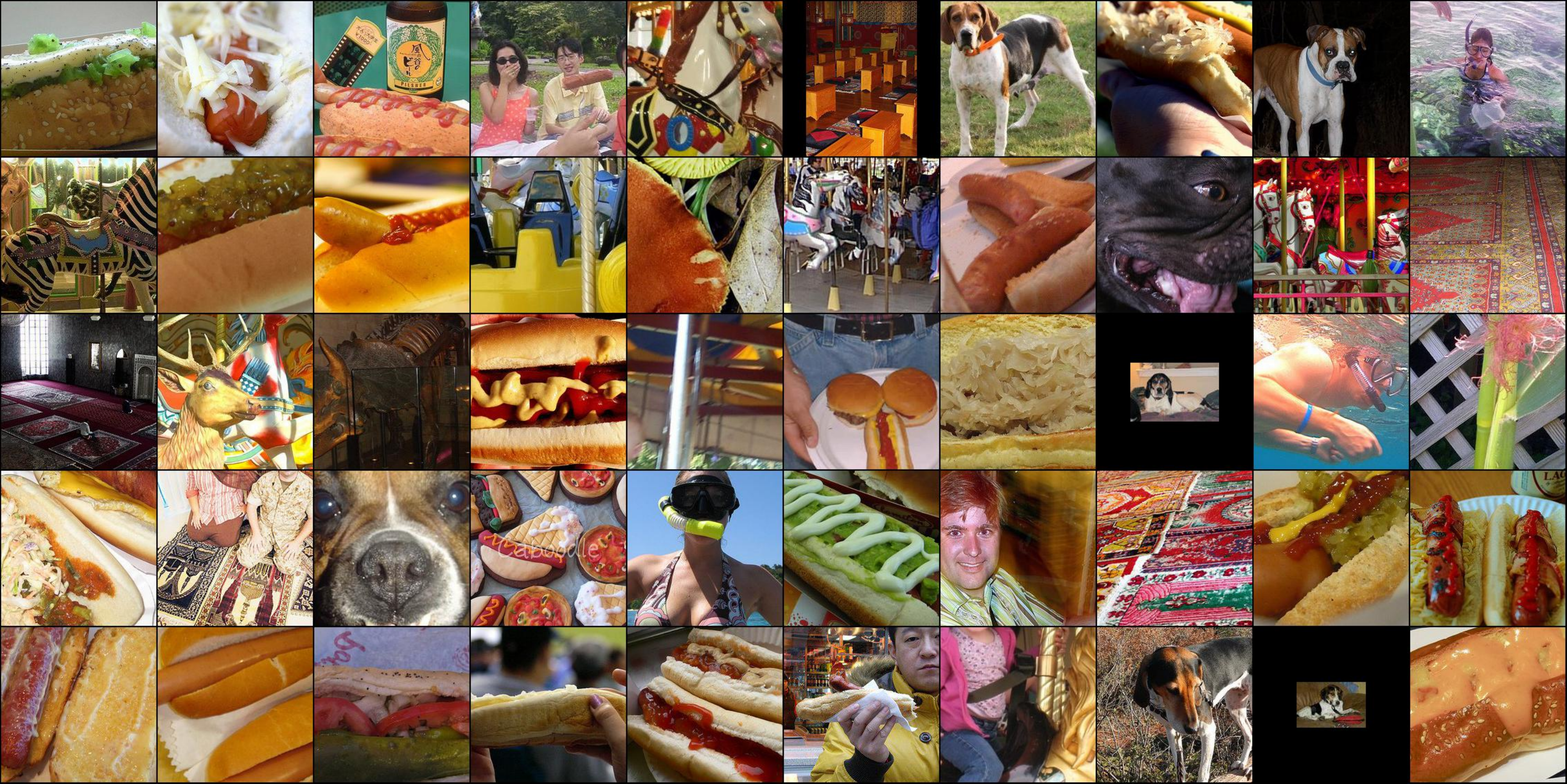} \\
    \end{tabular}
    \caption{Results of related base algorithm in a 5-way 5-shot scenario. Each column represents a different novel class. The top row shows the 5 novel instances, while the bottom row shows 60 randomly selected related base instances with $B=10$.}
    \label{fig:related-base}
\end{figure}

We develop a simple, yet effective procedure to select a set of base categories related to a novel category. Our method associates $B$ base categories to each novel class. After training $c(f(\cdot|\theta)|\mathbf{W})$ on $\mathcal{X}^b$, we first fine-tune $c(\cdot|\mathbf{W})$ on $\mathcal{X}^n$ while keeping $\theta$ fixed. Then, we define $\mathbf{M}\in\mathbb{R}^{K^b \times K^n}$ as a base-novel similarity matrix, where $K^b$ and $K^n$ are respectively the number of classes in $\mathcal{X}^b$ and $\mathcal{X}^n$. An element $m_{i,j}$ of the matrix $\mathbf{M}$ corresponds to the ratio of examples associated to the $i$-{th} base class that are classified as the $j$-{th} novel class: 
\begin{equation}
    \begin{split} 
        m_{i,j} &= \frac{1}{|\mathcal{X}_i^b|} \sum_{(\mathbf{x}_l^b,\cdot)\in\mathcal{X}_i^b} 
        \mathbb{I} \left[ j = \argmax_{k=1}^{K^n} \left(c_k(f(\mathbf{x}_l^b|\theta)\,|\,\mathbf{W})  \right) \right],
    \end{split}
\end{equation}
where $c_k(f(\mathbf{x}|\theta)|\mathbf{W})$ is the classifier output $c(\cdot|\mathbf{W})$ for class $k$. Then, the $B$ base classes with the highest score for a given novel class are kept as the related base for that class. Fig.~\ref{fig:related-base} illustrates example results obtained with this method in a 5-shot, 5-way scenario.

\subsection{Centroid associative alignment} 
\label{sec:alignment-centroid} 

Let us assume the set of instances $\mathcal{X}_i^n$ belonging to the $i$-th novel class $i\in\mathcal{Y}^n$, $\mathcal{X}_i^n=\{(\mathbf{x}^n_j,y_j^n)\in\mathcal{X}^n\,|\,y^n_j=i\}$, and the set of related base examples $\mathcal{X}_i^{rb}$ belonging to the same novel class $i$ according to the $g(\cdot|\mathbf{M})$ mapping function, $\mathcal{X}_i^{rb} = \{(\mathbf{x}^b_j,y^b_j)\in\mathcal{X}^{rb}\,|\,g(y_j|\mathbf{M})=i \}$. The function $g(y_j|\mathbf{M})\,:\,\mathcal{Y}^b\to\mathcal{Y}^n$ maps base class labels to the novel ones according to the similarity matrix $\mathbf{M}$. We wish to find an alignment transformation for matching probability densities $p(f(\mathbf{x}_{i,k}^n\,|\,\theta))$ and $p(f(\mathbf{x}_{i,l}^{rb}\,|\,\theta))$. Here, $\mathbf{x}_{i,k}^n$ is the $k$-th element from class $i$ in the novel set, and $\mathbf{x}_{i,l}^{rb}$ is the $l$-th element from class $i$ in the related base set. This approach has the added benefit of allowing the fine-tuning of all of the model parameters $\theta$ and $\mathbf{W}$ with a reduced level of overfitting.

\begin{figure*}[!t]
\begin{minipage}[t]{6.3cm}
  \ \\ 
     \begin{algorithm}[H]
    \SetAlgoLined
        \textbf{Input:} pre-trained model $c(f(\cdot|{\theta})|\mathbf{W})$, 
        novel class $\mathcal{X}^n$, related base set $\mathcal{X}^{rb}$.\\
        \textbf{Output:} fine-tuned $c(f(\cdot|{\theta})|\mathbf{W})$. \\ 
        \While{not done}
        { 
            $\widetilde{\mathcal{X}}^n \leftarrow \text{sample a batch from } \mathcal{X}^n$ \\
            $\widetilde{\mathcal{X}}^{rb} \leftarrow \text{sample a batch from } \mathcal{X}^{rb}$  \\ 
            \ \\
            evaluate $\mathcal{L}_\mathrm{ca}(\widetilde{\mathcal{X}}^{n},\widetilde{\mathcal{X}}^{rb})$, (eq.~\ref{centroid_loss_loss}) \\
            $\theta \leftarrow \theta - \eta_\mathrm{ca} \nabla_{\theta} \mathcal{L}_\mathrm{ca}(\widetilde{\mathcal{X}}^{n},\widetilde{\mathcal{X}}^{rb})$\\[0.5em]
    
            evaluate $\mathcal{L}_\mathrm{clf}(\widetilde{\mathcal{X}}^{rb})$, (eq.~\ref{eq:arcmax}) \\
            $\mathbf{W} \leftarrow \mathbf{W} - \eta_\mathrm{clf} \nabla_{\mathbf{W}} \mathcal{L}_\mathrm{clf}(\widetilde{\mathcal{X}}^{rb})$\\ 
            evaluate $\mathcal{L}_\mathrm{clf}(\widetilde{\mathcal{X}}^{n})$, (eq.~\ref{eq:arcmax}) \\
            $\mathbf{W} \leftarrow \mathbf{W} - \eta_\mathrm{clf} \nabla_{\mathbf{W}} \mathcal{L}_\mathrm{clf}(\widetilde{\mathcal{X}}^{n})$ \\ 
            $\theta \leftarrow \theta - \eta_\mathrm{clf} \nabla_{\theta} \mathcal{L}_\mathrm{clf}(\widetilde{\mathcal{X}}^{n})$ \\ 
        }
        \caption{\newline Centroid alignment.}
        \label{alg:centroid}
    \end{algorithm}
\end{minipage}%
~~
\begin{minipage}[t]{5.7cm} \ \\
  \centering
  \includegraphics[width=.62\linewidth]{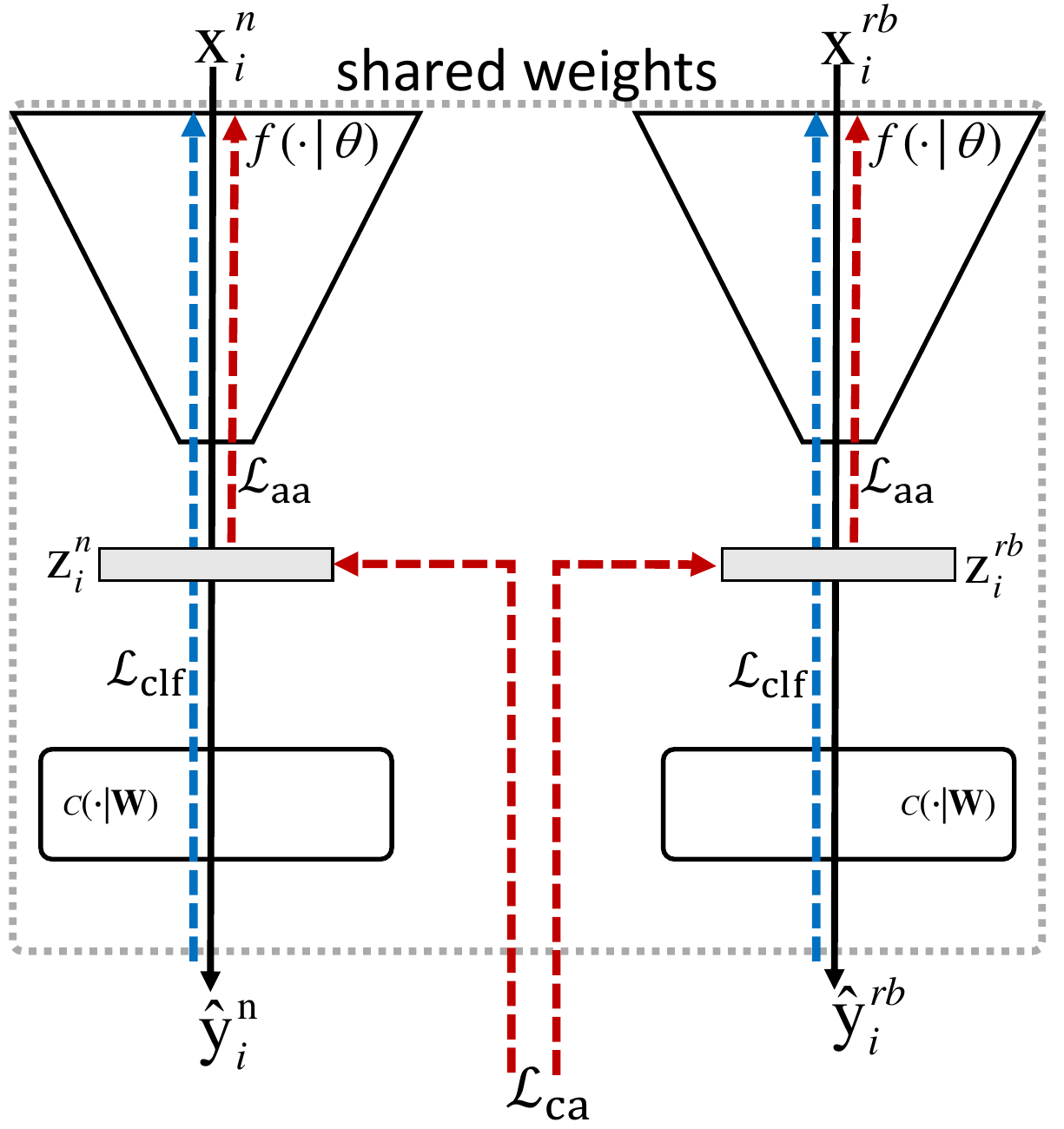}
        \caption{Schematic overview of our centroid alignment. The feature learner $f(\cdot|\theta)$ takes an example from novel category $\mathbf{x}^n$ and an example related base $\mathbf{x}_i^{rb}$. A Euclidean centroid based alignment loss $\mathcal{L}_\mathrm{ca}$ (red arrow) aligns the encoded $\mathbf{x}_i^n$ and $\mathbf{x}^{rb}_i$. Blue arrows represent classification loss $\mathcal{L}_\mathrm{clf}$.} 
        \label{fig:centroid-align} 
\end{minipage}
\end{figure*}

We propose a metric-based centroid distribution alignment strategy. The idea is to enforce intra-class compactness during the alignment process. Specifically, we explicitly push the training examples from the $i$-th novel class $\mathcal{X}_i^n$ 
towards the centroid of their related examples $\mathcal{X}_i^{rb}$
in feature space. 
The centroid $\bm{\mu}_i$ of $\mathcal{X}_i^{rb}$ is computed by
\begin{equation}
\bm{\mu}_i = \frac{1}{|\mathcal{X}_i^{rb}|} \sum_{(\mathbf{x}_j,\cdot)\in\mathcal{X}_i^{rb}} f(\mathbf{x}_j|\theta) \,,
\end{equation}
where $N^{n}$ and $N^{rb}$ are the number of examples in $\mathcal{X}^{n}$ and $\mathcal{X}^{rb}$, respectively. This allows the definition of the centroid alignment loss as
\begin{equation} 
      \mathcal{L}_\mathrm{ca}(\mathcal{X}^n) =  
    -\frac{1}{N^n N^{rb}} \sum_{i=1}^{K^n} \sum_{(\mathbf{x}_j,\cdot) \in \mathcal{X}_i^n} \log 
     \frac{ \exp[-\|f(\mathbf{x}_j|\theta) - \bm{\mu}_i\|^2_2]}
     {\sum\nolimits_{k=1}^{K^n} \exp[-\|f( \mathbf{x}_j | \theta) - \bm{\mu}_{k}\|^2_2]} \,.
     \label{centroid_loss_loss} 
\end{equation}
Our alignment strategy bears similarities to \cite{snell2017prototypical} which also uses eq.~\ref{centroid_loss_loss} in a meta-learning framework. In our case, we use that same equation to match distributions. 
Fig.~\ref{fig:centroid-align} illustrates our proposed centroid alignment, and algorithm~\ref{alg:centroid} presents the overall procedure. First, we update the parameters of the feature extraction network $f(\cdot|\theta)$ using eq.~\ref{centroid_loss_loss}. Second, the entire network is updated using a classification loss $\mathcal{L}_\mathrm{clf}$ (defined in sec.~\ref{sec:baseline}).

%
%

%
%
%

%

%
\begin{figure*}[!t]
\begin{minipage}[t]{6.3cm}
  \ \\  
    \begin{algorithm}[H]
    \SetAlgoLined
        \textbf{Input:} pre-trained model $c(f(\cdot|{\theta})|\mathbf{W})$, 
        novel class $\mathcal{X}^n$, related base set $\mathcal{X}^{rb}$.\\
        \textbf{Output:} fine-tuned $c(f(\cdot|{\theta})|\mathbf{W})$. \\ 
        \While{not done}
        {
            $\widetilde{\mathcal{X}}^n \leftarrow \text{sample a batch from } \mathcal{X}^n$ \\
            $\widetilde{\mathcal{X}}^{rb} \leftarrow \text{sample a batch from } \mathcal{X}^{rb}$  \\[0.5em]
            \For{i = 0,\ldots,$n_\mathrm{critic}$}
            {   
                evaluate $\mathcal{L}_{h}(\widetilde{\mathcal{X}}^{n},\widetilde{\mathcal{X}}^{rb})$, (eq.~\ref{eq:adv-critic}) \\  
                $\triangleright$ update critic: \\ 
                $\phi \leftarrow \phi + \eta_{h} \nabla_{\phi} \mathcal{L}_{h}(\widetilde{\mathcal{X}}^{n},\widetilde{\mathcal{X}}^{rb})$ \\
                $\phi \leftarrow \mathrm{clip}(\phi, -0.01, 0.01)$
            }
            \ \\
            evaluate $\mathcal{L}_\mathrm{aa}(\widetilde{\mathcal{X}}^{n})$, (eq.~\ref{eq:adv-feature}) \\
            $\theta \leftarrow \theta - \eta_\mathrm{aa} \nabla_{\theta} \mathcal{L}_\mathrm{aa}(\widetilde{\mathcal{X}}^{n})$\\
    
            evaluate $\mathcal{L}_\mathrm{clf}(\widetilde{\mathcal{X}}^{rb})$, (eq.~\ref{eq:arcmax}) \\
            $\mathbf{W} \leftarrow \mathbf{W} - \eta_\mathrm{clf} \nabla_{\mathbf{W}} \mathcal{L}_\mathrm{clf}(\widetilde{\mathcal{X}}^{rb})$\\
            evaluate $\mathcal{L}_\mathrm{clf}(\widetilde{\mathcal{X}}^{n})$, (eq.~\ref{eq:arcmax}) \\
            $\mathbf{W} \leftarrow \mathbf{W} - \eta_\mathrm{clf} \nabla_{\mathbf{W}} \mathcal{L}_\mathrm{clf}(\widetilde{\mathcal{X}}^{n})$ \\
            $\theta \leftarrow \theta - \eta_\mathrm{clf} \nabla_{\theta} \mathcal{L}_\mathrm{clf}(\widetilde{\mathcal{X}}^{n})$ \\ 
        } 
        \caption{\newline Adversarial alignment}
        \label{alg:adversarial}
    \end{algorithm}
\end{minipage}%
~~
\begin{minipage}[t]{5.7cm} \ \\
  \centering
  \includegraphics[width=.64\linewidth]{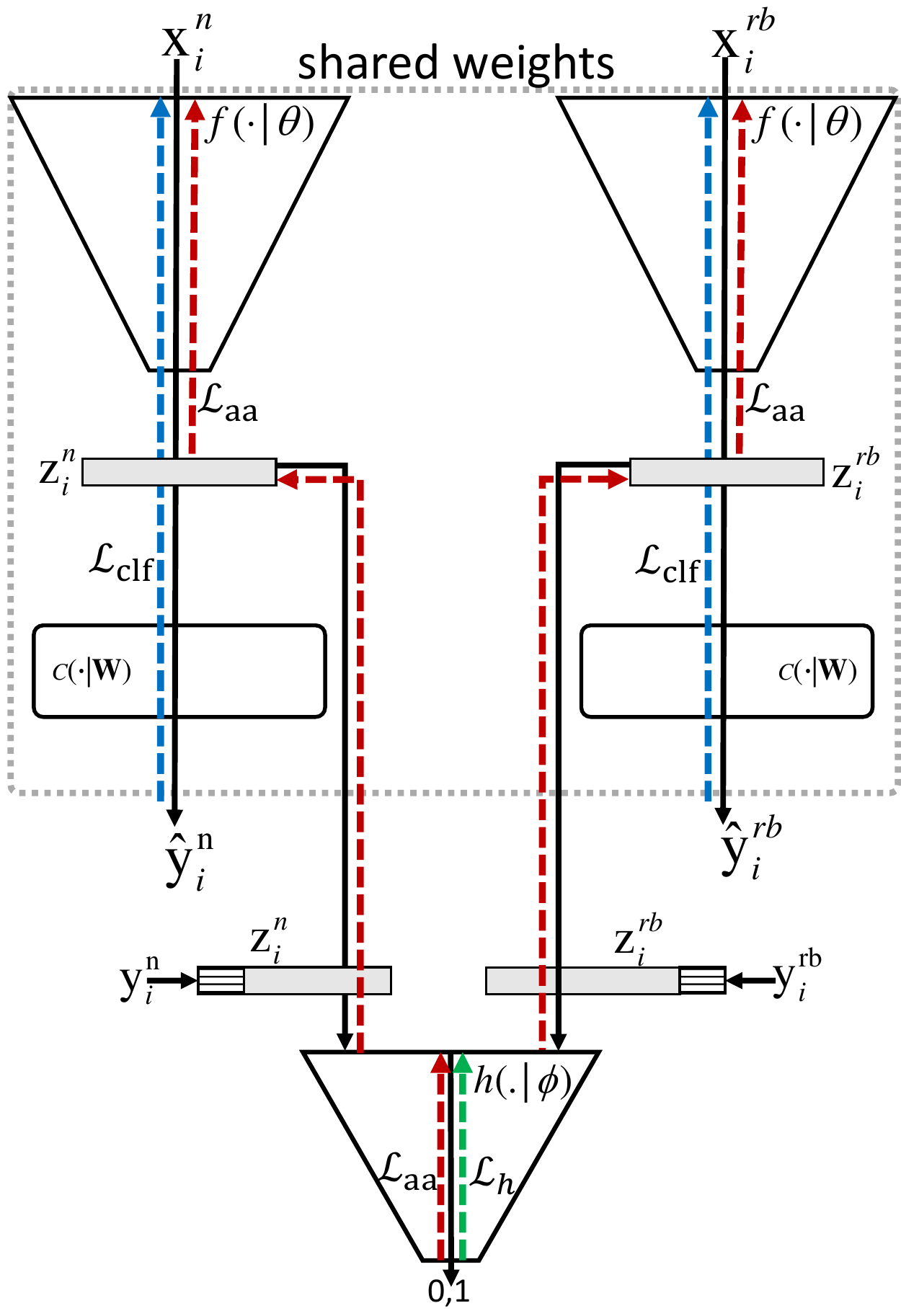} 
        \caption{ Overview of our adversarial alignment. The feature learner $f(\cdot|\theta)$ takes an image $\mathbf{x}^{n}_i$ from the $i$-{th} novel class and an example $\mathbf{x}^{rb}_i$ of the related base. The critic $h(\cdot|\phi)$ takes the feature vectors and the one-hot class label vector. Green, red and blue arrows present the critic $\mathcal{L}_{h}$, adversarial $\mathcal{L}_{aa}$ and classification $\mathcal{L}_\mathrm{clf}$ losses respectively. }
        \label{fig:adversarial-align}
\end{minipage}
\end{figure*}

\subsection{Adversarial associative alignment}
\label{sec:alignment-adversarial}
As an alternative associative alignment strategy, and inspired by WGAN~\cite{arjovsky2017wasserstein}, we experiment with training the encoder $f(\cdot|\theta)$ to perform adversarial alignment using a conditioned critic network $h(\cdot|{\phi})$ based on Wasserstein-1 distance between two probability densities $p_{x}$ and $p_{y}$:
\begin{equation}
    D(p_{x}, p_{y}) = \sup_{\| h \|_L \leq 1} \mathbb{E}_{x \sim p_{x}} [h(x)] - \mathbb{E}_{x\sim p_{y}} [h(x)] \,,
\end{equation}
where $\sup$ is the supremum, and $h$ is a 1-Lipschitz function.
Similarly to Arjovsky \emph{et al.}~\cite{arjovsky2017wasserstein}, we use a parameterized critic network $h(\cdot|\phi)$ conditioned by the concatenation of the feature embedding of either $\mathbf{x}^n_i$ or $\mathbf{x}_j^{rb}$, along with the corresponding label $y_i^n$ encoded as a one-hot vector. Conditioning $h(\cdot|\phi)$ helps the critic in matching novel categories and their corresponding related base categories. The critic $h(\cdot|\phi)$ is trained with loss
\begin{align}
    \label{eq:adv-critic}
    \mathcal{L}_{h}(\mathcal{X}^{n},\mathcal{X}^{rb}) = & ~ \frac{1}{N^{rb}} \sum_{(\mathbf{x}^{rb}_i,y_i^{rb})\in\mathcal{X}^{rb}} h \left([f(\mathbf{x}_i^{rb} | \theta) \;  y_i^{rb}]\,|\,\phi \right) \nonumber \\
    & ~ - \frac{1}{N^{n}} \sum_{(\mathbf{x}_i^n,y_i^n)\in\mathcal{X}^{n}}  h \left( [f(\mathbf{x}_i^n | \theta) \; y_i^n]\,|\,\phi \right) \,,
\end{align}
where, $[\cdot]$ is the concatenation operator. Then, the encoder parameters $\theta$ are updated using
\begin{equation}
\label{eq:adv-feature}
    \mathcal{L}_\mathrm{aa}(\mathcal{X}^{n})  = \frac{1}{K^n} \sum_{(\mathbf{x}_i^n,y_i^n) \in \mathcal{X}^{n}} h \left( [f(\mathbf{x}^n_i | \theta) \;  y_i^n] | \phi \right) \,.
\end{equation} 
Algorithm~\ref{alg:adversarial} summarizes our adversarial alignment method. First, we perform the parameter update of critic $h(\cdot|\phi)$ using eq.~\ref{eq:adv-critic}. Similar to WGAN~\cite{arjovsky2017wasserstein}, we perform $n_\mathrm{critic}$ iterations to optimize $h$, before updating $f(\cdot|\theta)$ using eq.~\ref{eq:adv-feature}. Finally, the entire network is updated by a classification loss $\mathcal{L}_\mathrm{clf}$ (defined in sec.~\ref{sec:baseline}).

\section{Establishing a strong baseline} 
\label{sec:baseline}
Before evaluating our alignment strategies in sec.~\ref{sec:experiments}, we first establish a strong baseline for comparison by following the recent literature. In particular, we build on the work of Chen \emph{et al.}~\cite{chen2019closer} but incorporate a different loss function and episodic early stopping on the pre-training stage.

\subsection{Classification loss functions} 
\label{sec:arcmax}
Deng \emph{et al.}~\cite{deng2018arcface} have shown that an additive angular margin (``arcmax'' hereafter) outperforms other metric learning algorithms for face recognition. The arcmax has a metric learning property since it enforces a geodesic distance margin penalty on the normalized hypersphere, which we think can be beneficial for few-shot classification by helping keep class clusters compact and well-separated. 

Let $\mathbf{z}$ 
be the representation of $\mathbf{x}$ in feature space. As per \cite{deng2018arcface}, we transform the logit as $\mathbf{w}_j^\top \mathbf{z} = \|\mathbf{w}_j \| \|\mathbf{z} \| \cos \varphi_j$, where $\varphi_j$ is the angle between $\mathbf{z}$ and $\mathbf{w}_j$, the $j$-{th} column in the weight matrix $\mathbf{W}$. Each weight $\|\mathbf{w}_j\| = 1$ by $l_2$ normalization. Arcmax adds an angular margin $m$ to the distributed examples on a hypersphere:  
\begin{equation}
\label{eq:arcmax} 
  \mathcal{L}_{\mathrm{clf}} = 
 -\frac{1}{N} \sum_{i=1}^N \log \frac{\exp(s\cos (\varphi_{y_i}+m))}{\exp(s \cos (\varphi_{y_i}+m)) + \sum\limits_{\forall j\neq y_i} \exp(s \cos \varphi_j)} \,,
\end{equation} 
where $s$ is the radius of the hypersphere on which $\mathbf{z}$ is distributed, 
$N$ the number of examples, and $m$ and $s$ are hyperparameters (see sec.~\ref{sec:datasets-implementation}). The overall goal of the margin is to enforce inter-class discrepancy and intra-class compactness.
\subsection{Episodic early stopping} 
\label{sec:early-stopping}
A fixed number of epochs in the pre-training stage has been commonly used (e.g., \cite{chen2019closer,finn2017model,snell2017prototypical,vinyals2016matching}), but this might hamper performance in the fine-tuning stage. Using validation error, we observe the necessity of early-stopping in pre-training phase (see supp. mat. for a validation error plot). 
%
%
We thus make the use of episodic early stopping using validation set at pre-training time,  specifically by stopping the training when the mean accuracy 
over a window of recent epochs starts to decrease. The best model in the window is selected as the final result.

\section{Experimental validation} 
\label{sec:experiments}
In the following, we are conducting an experimental evaluation and comparison of the proposed associative alignment strategies for few-shot learning. First, we introduce the datasets used and evaluate the strong baseline from sec.~\ref{sec:baseline}. 
\subsection{Datasets and implementation details} 
\label{sec:datasets-implementation}
\paragraph{Datasets} We present experiments on four benchmarks: \textit{mini}-ImageNet~\cite{vinyals2016matching}, tieredImageNet~\cite{ren2018meta}, and FC100~\cite{oreshkin2018tadam} for generic object recognition; and CUB-200-2011 (CUB)~\cite{wah2011caltech} for fine-grained image classification. 
\textit{mini}-ImageNet is a subset of the ImageNet ILSVRC-12 dataset~\cite{russakovsky2015imagenet} containing 100 categories and 600 examples per class. We used the same splits as Ravi and Larochelle~\cite{ravi2016optimization}, where 64, 16, and 20 classes are used for the base, validation, and novel classes, respectively. As a larger benchmark, the tieredImageNet~\cite{ren2018meta} is also a subset of ImageNet ILSVRC-12 dataset~\cite{russakovsky2015imagenet}, this time with 351 base, 97 validation, and 160 novel classes respectively. Derived from CIFAR-100~\cite{krizhevsky2009cifar}, the FC100 dataset~\cite{oreshkin2018tadam} contains 100 classes grouped into 20 superclasses to minimize class overlap. Base, validation and novel splits contain 60, 20, 20 classes belonging to 12, 5, and 5 superclasses, respectively.
The CUB dataset~\cite{wah2011caltech} contains 11,788 images from 200 bird categories. We used the same splits as Hilliard \emph{et al.}~\cite{hilliard2018few} using 100, 50, and 50 classes for the base, validation, and novel classes, respectively. 
\paragraph{Network architectures} We experiment with three backbones for the feature learner $f(\cdot|\theta)$: 1) a 4-layer convolutional network  (``Conv4'') with input image resolution of $84 \times 84$, similar to \cite{finn2017model,ravi2016optimization,snell2017prototypical}; 2) a ResNet-18~\cite{he2016deep} with input size of $224 \times 224$; and 3) a 28-layers Wide Residual Network (``WRN-28-10'')~\cite{sergey2016wide} with input size of $80 \times 80$ in 3 steps of dimension reduction. 
We use a single hidden layer MLP of 1024 dimensions as the critic network $h(\cdot|\phi)$ (c.f. sec.~\ref{sec:alignment-adversarial}).
%
%
\paragraph{Implementation details} Recall from sec.~\ref{sec:preliminaries} that training consists of two stages: 1) pre-training using base categories $\mathcal{X}^b$; and 2) fine-tuning on novel categories $\mathcal{X}^n$. For pre-training, we use the early stopping algorithm from sec.~\ref{sec:early-stopping} with a window size of 50. Standard data augmentation approaches (i.e., color jitter, random crops, and left-right flips as in \cite{chen2019closer}) have been employed, and the Adam algorithm with a learning rate of $10^{-3}$ and batch size of 64 is used for both pre-training and fine-tuning. The arcmax loss (eq.~\ref{eq:arcmax}) is configured with $s = 20$ and $m=0.1$ which are set by cross validation.
In the fine-tuning stage, episodes are defined by randomly selecting $N=5$ classes from the novel categories $\mathcal{X}^n$. $k$ examples for each category are subsequently sampled ($k=1$ and $k=5$ in our experiments). As in Chen \emph{et al.}~\cite{chen2019closer}, no standard data augmentation was used in this stage.  We used episodic cross-validation to find $s$ and $m$ with a fixed encoder. More specifically, $(s,m)$ were found to be $(5, 0.1)$ for the Conv4 and $(5, 0.01)$ for the WRN-28-10 and ResNet-18 backbones. The learning rate for Adam was set to $10^{-3}$ and $10^{-5}$ for the centroid and adversarial alignments respectively. Similarly to \cite{arjovsky2017wasserstein}, 5 iterations (inner loop of algorithm~\ref{alg:adversarial}) were used to train the critic $h(\cdot|\phi)$. We fix the number of related base categories as $B=10$ (see supp. mat. for an ablation study on $B$). For this reason, we used a relatively large number of categories (50 classes out of the 64 available in \textit{mini}-ImageNet).


\begin{table*}[t]
\centering
\caption{Preliminary evaluation using \textit{mini}-ImageNet and CUB, presenting 5-way classification accuracy using the Conv4 backbone, with $\pm$ indicating the 95\% confidence intervals over 600 episodes. The best result is boldfaced, while the best result \emph{prior to this work} is highlighted in blue. Throughout this paper, ``--'' indicates when a paper does not report results in the corresponding scenario.} 
    \begin{tabular}{ rlccccccc} 
        \toprule
        
        & 
        & \multicolumn{2}{c}{\textit{mini}-ImageNet}  
        &
        & \multicolumn{2}{c}{CUB}   
        \\ 
        
        & \textbf{Method}  
        & \textbf{1-shot}  & \textbf{5-shot}  
        &
        & \textbf{1-shot}  & \textbf{5-shot} 
        \\ 
           \midrule
        \multirow{5}{*}{\rotatebox{90}{meta learning}}  
                     & Meta-LSTM~\cite{ravi2016optimization}   
                     & 43.44\scriptsize{ $\pm$ 0.77}   & 55.31\scriptsize{ $\pm$ 0.71}  
                     &
                     & -- &--   \\

                     & MatchingNet$^\ddag$~\cite{vinyals2016matching}           
                     & 43.56\scriptsize{ $\pm$ 0.84}  & 55.31\scriptsize{ $\pm$ 0.73}
                     &
                     & 60.52\scriptsize{ $\pm$ 0.88}      & 75.29\scriptsize{ $\pm$ 0.75}  \\ 
                     
                     & ProtoNet$^\ddag$~\cite{snell2017prototypical}           
                     & 49.42\scriptsize{ $\pm$ 0.78}      & 68.20\scriptsize{ $\pm$ 0.66}
                     &
                     & 51.31\scriptsize{ $\pm$ 0.91}      & 70.77\scriptsize{ $\pm$ 0.69}   \\
                     
                     & MAML$^\ddag$~\cite{finn2018probabilistic}              
                     & 48.07\scriptsize{ $\pm$ 1.75}      & 63.15\scriptsize{ $\pm$ 0.91}
                     &
                     & 55.92\scriptsize{ $\pm$ 0.95}      & 72.09\scriptsize{ $\pm$ 0.76}         \\
                     
                     & RelationNet$^\ddag$~\cite{sung2018learning}    
                     & 50.44\scriptsize{ $\pm$ 0.82}      & 65.32\scriptsize{ $\pm$ 0.70}
                     &
                     & {\color{blue}62.45\scriptsize{ $\pm$ 0.98}}      & 76.11\scriptsize{ $\pm$ 0.69}    \\ 
                     
                     \midrule
        \multirow{6}{*}{\rotatebox{90}{tr. learning}} 
        
                    & softmax$^\dag$                  & 46.40\scriptsize{ $\pm$ 0.72}      & 64.37\scriptsize{ $\pm$ 0.59}  
                    &
                    & 47.12\scriptsize{ $\pm$ 0.74}                & 64.16\scriptsize{ $\pm$ 0.71}
                    \\

                    & softmax$^{\dag\diamond}$        & 46.99\scriptsize{ $\pm$  0.73}       & 65.33\scriptsize{ $\pm$ 0.60}    
                    &
                    & 45.68\scriptsize{ $\pm$ 0.86}                & 66.94\scriptsize{ $\pm$ 0.84} 
                    \\

                    & cosmax$^\dag$                  & 50.92\scriptsize{ $\pm$ 0.76}          & 67.29\scriptsize{ $\pm$ 0.59}     
                    &
                    & 60.53\scriptsize{ $\pm$ 0.83}      & 79.34\scriptsize{ $\pm$ 0.61}
                    \\
                     
                    & cosmax$^{\dag\diamond}$        & {\color{blue}52.04\scriptsize{ $\pm$ 0.82}}      & {\color{blue}68.47\scriptsize{ $\pm$ 0.60}}    
                    &
                    & 60.66\scriptsize{ $\pm$ 1.04}      & {\color{blue}79.79\scriptsize{ $\pm$ 0.75}}    
                    \\

                    & our baseline (sec.~\ref{sec:baseline})   & 51.90\scriptsize{ $\pm$ 0.79}    & 69.07\scriptsize{ $\pm$ 0.62} 
                    &
                    & 60.85\scriptsize{ $\pm$ 1.07}      & 79.74\scriptsize{ $\pm$ 0.64}
                    \\
                    \midrule
        \multirow{2}{*}{\rotatebox{90}{align.}} 
                    & adversarial  
                    & 52.13\scriptsize{ $\pm$ 0.99}      & 70.78\scriptsize{ $\pm$ 0.60}
                    &
                    & \textbf{63.30}\scriptsize{ $\pm$ 0.94}                 & \textbf{81.35}\scriptsize{ $\pm$ 0.67}    \\

                    & centroid           
                    & \textbf{53.14}\scriptsize{ $\pm$ 1.06}      & \textbf{71.45}\scriptsize{ $\pm$ 0.72}
                    &
                    & 62.71\scriptsize{ $\pm$ 0.88}               & 80.48\scriptsize{ $\pm$ 0.81}  \\
                    \bottomrule
    \end{tabular}
    \\
    $^{\dag}$ our implementation \quad   $^{\diamond}$  with early stopping \quad 
    $^{\ddag}$ implementation from \cite{chen2019closer} for CUB 
    \label{tab:baselines}
\end{table*}

\subsection{\textit{mini}-ImageNet and CUB with a shallow Conv4 backbone}
\label{sec:baseline-evaluation}
We first evaluate the new baseline presented in sec.~\ref{sec:baseline} and our associative alignment strategies using the Conv4 backbone on the \textit{mini}-ImageNet (see supp. mat. for evaluations in higher number of ways) and CUB datasets, with corresponding results presented in table~\ref{tab:baselines}. 
%
%
We note that arcmax with early stopping improves on using cosmax and softmax with and without early stopping for both the 1- and 5-shot scenarios, on both the \textit{mini}-ImageNet and CUB datasets. 
We followed the same dataset split configuration, network architecture, and implementation details given in \cite{chen2019closer} for our testing. 
Our centroid associative alignment outperforms the state of the art in all the experiments, with gains of 1.24\% and 2.38\% in 1- and 5-shot over our baseline on \textit{mini}-ImageNet. For CUB, the adversarial alignment provides an additional gain of 0.6\% and 0.87\% over the centroid one. 
%
%
\begin{table*}[!t]
\small{
\centering
\caption{\textit{mini}-ImageNet and tieredImageNet results using ResNet-18 and WRN-28-10 backbones. 
$\pm$ denotes the $95\%$ confidence intervals over 600 episodes.}
    \begin{tabular}{ rlccccccc} 
        \toprule
        
        & 
        & \multicolumn{2}{c}{\textit{mini}-ImageNet}  
        &
        & \multicolumn{2}{c}{tieredImageNet}   
        \\ 
        
        & \textbf{Method}  
        & \textbf{1-shot}  & \textbf{5-shot}  
        &
        & \textbf{1-shot}  & \textbf{5-shot} 
        \\ 
        \midrule
            \multirow{11}{*}{\rotatebox{90}{ResNet-18} \quad } 
            
                    & TADAM~\cite{oreshkin2018tadam}        
                    & 58.50\scriptsize{ $\pm$ 0.30}                          & 76.70\scriptsize{ $\pm$ 0.30} 
                    &
                    & --                           & --  
                    \\
                    
                    & ProtoNet$^{\ddag}$~\cite{snell2017prototypical}        
                    & 54.16\scriptsize{ $\pm$ 0.82}            & 73.68\scriptsize{ $\pm$ 0.65} 
                    &
                    & 61.23\scriptsize{ $\pm$ 0.77}            & 80.00\scriptsize{ $\pm$ 0.55} 
                    \\
                    
                    & SNAIL~\cite{mishra2017simple}       
                    & 55.71\scriptsize{ $\pm$ 0.99}           & 68.88\scriptsize{ $\pm$ 0.92} 
                    &
                    & --                                & --  
                    \\

                    & IDeMe-Net~\cite{Chen_2019_CVPR}      
                    & 59.14\scriptsize{ $\pm$ 0.86}           & 74.63\scriptsize{ $\pm$ 0.74} 
                    &
                    & --                           & --  
                    \\
                    
                    & Activation to Param.~\cite{Qiao_2018_CVPR}        
                    & 59.60\scriptsize{ $\pm$ 0.41}            & 73.74\scriptsize{ $\pm$ 0.19} 
                    &
                    & --                                  & --  
                    \\
                    
                    & MTL~\cite{sun2019meta}       
                    & 61.20\scriptsize{ $\pm$ 1.80}           & 75.50\scriptsize{ $\pm$ 0.80} 
                    &
                    & --                           & --  
                    \\

                    & TapNet~\cite{yoon2019tapnet}        
                    & 61.65\scriptsize{ $\pm$ 0.15}            & 76.36\scriptsize{ $\pm$ 0.10} 
                    &
                    & 63.08\scriptsize{ $\pm$ 0.15}            & 80.26\scriptsize{ $\pm$ 0.12} 
                    \\

                    & VariationalFSL~\cite{zhang2019variational}       
                    & 61.23\scriptsize{ $\pm$ 0.26}           & 77.69\scriptsize{ $\pm$ 0.17} 
                    &
                    & --                           & --  
                    \\

                    & MetaOptNet$^*$~\cite{lee2019meta}        
                    & {\color{blue}\textbf{62.64}\scriptsize{ $\pm$ 0.61}}                & {\color{blue}78.63\scriptsize{ $\pm$ 0.46}} 
                    &
                    & {\color{blue}65.99\scriptsize{ $\pm$ 0.72}}                & {\color{blue}81.56\scriptsize{ $\pm$ 0.53}} 
                    \\

                    \cmidrule(lr){2-7}

                    & our baseline (sec.~\ref{sec:baseline})        
                    & 58.07\scriptsize{ $\pm$ 0.82}            & 76.62\scriptsize{ $\pm$ 0.58} 
                    &
                    & 65.08\scriptsize{ $\pm$ 0.19}            & 83.67\scriptsize{ $\pm$ 0.51}
                    \\

                    & adversarial alignment      
                    & 58.84\scriptsize{ $\pm$ 0.77}                          & 77.92 \scriptsize{ $\pm$ 0.82}
                    &
                    & 66.44\scriptsize{ $\pm$ 0.61}                          & 85.12\scriptsize{ $\pm$ 0.53} 
                    \\ 
                    
                    & centroid alignment            
                    & 59.88\scriptsize{ $\pm$ 0.67}           & \textbf{80.35}\scriptsize{ $\pm$ 0.73} 
                    &
                    & \textbf{69.29}\scriptsize{ $\pm$ 0.56}           & \textbf{85.97}\scriptsize{ $\pm$ 0.49} 
                    \\

            \toprule       
            \multirow{9}{*}{\rotatebox{90}{ WRN-28-10} \quad }

                    & LEO~\cite{rusu2018meta}  
                    & 61.76\scriptsize{ $\pm$ 0.08}           & 77.59\scriptsize{ $\pm$ 0.12} 
                    &
                    & 66.33\scriptsize{ $\pm$ 0.09}           & 81.44 \scriptsize{ $\pm$ 0.12} 
                    \\

                    & wDAE~\cite{gidaris2019generating}  
                    & 61.07\scriptsize{ $\pm$ 0.15}           & 76.75\scriptsize{ $\pm$ 0.11} 
                    &
                    & 68.18\scriptsize{ $\pm$ 0.16}           & 83.09\scriptsize{ $\pm$ 0.12} 
                    \\

                    & CC+rot~\cite{gidaris2019boosting}  
                    & 62.93\scriptsize{ $\pm$ 0.45}           & 79.87\scriptsize{ $\pm$ 0.33} 
                    &
                    & 70.53\scriptsize{ $\pm$ 0.51}           & 84.98\scriptsize{ $\pm$ 0.36} 
                    \\

                    & Robust-dist++~\cite{rusu2018meta}  
                    & 63.28\scriptsize{ $\pm$ 0.62}           & {\color{blue}81.17\scriptsize{ $\pm$ 0.43}} 
                    &
                    & --                                & --
                    \\

                    & Transductive-ft~\cite{dhillon2019baseline}  
                    & {\color{blue}65.73\scriptsize{ $\pm$ 0.68}}           & 78.40\scriptsize{ $\pm$ 0.52} 
                    &
                    & {\color{blue}73.34\scriptsize{ $\pm$ 0.71}}           & {\color{blue}85.50\scriptsize{ $\pm$ 0.50}} 
                    \\

                    \cmidrule(lr){2-7} 
                    
                    & our baseline (sec.~\ref{sec:baseline})        
                    & 63.28\scriptsize{ $\pm$0.71}            & 78.31\scriptsize{ $\pm$0.57} 
                    &
                    & 68.47\scriptsize{ $\pm$0.86}            & 84.11\scriptsize{ $\pm$0.65} 
                    \\
                    
                    & adversarial alignment      
                    & 64.79\scriptsize{ $\pm$0.93}            & 82.02\scriptsize{ $\pm$0.88}
                    &
                    & 73.87\scriptsize{ $\pm$0.76}                    & 84.95\scriptsize{ $\pm$0.59} 
                    \\ 
                    
                    & centroid alignment            
                    & \textbf{65.92}\scriptsize{ $\pm$ 0.60}            & \textbf{82.85}\scriptsize{ $\pm$ 0.55} 
                    &
                    & \textbf{74.40}\scriptsize{ $\pm$ 0.68}                          & \textbf{86.61}\scriptsize{ $\pm$0.59} 
                    \\
        \bottomrule
    \end{tabular}
    \\ 
    $^\ddag$ Results are from~\cite{chen2019closer} for {\textit{mini}-ImageNet} and from~\cite{lee2019meta} for tieredImageNet,
    * ResNet-12 
    \label{tab:ImageNet}
    }
\end{table*}

\subsection{\textit{mini}-ImageNet and tieredimageNet with deep backbones}
We now evaluate our proposed associative alignment on both the \textit{mini}-ImageNet and tieredimageNet datasets using two deep backbones: ResNet-18 and WRN-28-10. 
Table~\ref{tab:ImageNet} compares our proposed alignment methods with several approaches.

\paragraph{\textit{mini}-ImageNet} 
Our centroid associative alignment strategy achieves the best 1- and 5-shot classification tasks on both the ResNet-18 and WRN-28-10 backbones, with notable absolute accuracy improvements of 2.72\% and 1.68\% over MetaOptNet~\cite{lee2019meta} and Robust-dist++~\cite{dvornik2019diversity} respectively. The single case where a previous method achieves superior results is that of MetaOptNet,
which outperforms our method by 2.76$\%$ in 1-shot. For the WRN-28-10 backbone, we achieve similar results to Transductive-ft~\cite{dhillon2019baseline} for 1-shot, but outperform their method by 4.45\% in 5-shot.
%
Note that unlike IDeMe-Net~\cite{Chen_2019_CVPR}, SNAIL~\cite{mishra2017simple} and TADAM~\cite{oreshkin2018tadam}, which make use of extra modules, our method achieves significant improvements over these methods without any changes to the backbone. 

%
\paragraph{tieredImageNet} Table~\ref{tab:ImageNet} also shows that our centroid associative alignment outperforms the compared methods on tieredImageNet in both 1- and 5-shot scenarios. Notably, our centroid alignment results in a gain of 3.3$\%$ and 4.41$\%$ over MetaOptNet~\cite{lee2019meta} using the ResNet-18. 
Likewise, our centroid alignment gains 1.06$\%$ and 1.11$\%$ over the best of the compared methods using WRN-28-10.

\subsection{FC100 and CUB with a ResNet-18 backbone}
We present additional results on the FC100 and CUB datasets with a ResNet-18 backbone in table~\ref{tab:FC_CUB}. 
%
In FC100, our centroid alignment gains 0.73$\%$ and 2.14$\%$ over MTL~\cite{sun2019meta} in 1- and 5-shot respectively.
%
We also observe improvements in CUB with our associative alignment approaches, with the centroid alignment outperforming ProtoNet~\cite{snell2017prototypical} by 2.3\% in 1-shot and 1.2\% in 5-shot.
We outperform Robust-20~\cite{dvornik2019diversity}, an ensemble of 20 networks, by 4.03\% and 4.15\% on CUB.
\begin{table*}[!t]
\centering
\caption{Results on the FC100 and CUB dataset using ResNet-18 backbones. $\pm$ denotes the $95\%$ confidence intervals over 600 episodes. The best result is boldfaced, while the best result \emph{prior to this work} is highlighted in blue.} 
    \begin{tabular}{ rlccccccc} 
        \toprule
        
        & 
        & \multicolumn{2}{c}{FC100}  
        &
        & \multicolumn{2}{c}{CUB}   
        \\ 
        
        & \textbf{Method}  
        & \textbf{1-shot}  & \textbf{5-shot}  
        &
        & \textbf{1-shot}  & \textbf{5-shot} 
        \\ 
           \midrule 

                     & Robust-20~\cite{dvornik2019diversity}        
                     & --  & --    
                     &        
                     & 58.67\scriptsize{ $\pm$ 0.65}     & 75.62\scriptsize{ $\pm$ 0.48}     \\

                     & GNN-LFT~\cite{tseng2020cross}        
                     &--  &--    
                     &        
                     & 51.51\scriptsize{ $\pm$ 0.80}     & 73.11\scriptsize{ $\pm$ 0.68}     \\

                     & RelationNet$^\ddag$~\cite{sung2018learning}             
                     & --             & --       
                     &       
                     & 67.59\scriptsize{ $\pm$ 1.02}             & 82.75\scriptsize{ $\pm$ 0.58}     \\

                     & ProtoNet$^\ddag$~\cite{snell2017prototypical}           
                     & 40.5\scriptsize{ $\pm$ 0.6}       & 55.3\scriptsize{ $\pm$ 0.6}  
                     &       
                     & {\color{blue}71.88\scriptsize{ $\pm$ 0.91}}             & {\color{blue}87.42\scriptsize{ $\pm$ 0.48}}     \\
                     
                     
                     & TADAM~\cite{oreshkin2018tadam}                 
                     & 40.1\scriptsize{ $\pm$ 0.4}       & 56.1\scriptsize{ $\pm$ 0.4}    
                     &        
                     &--  &--         \\

                     & MetaOptNet$^\dag$~\cite{lee2019meta}                       
                     & 41.1\scriptsize{ $\pm$ 0.6}       & 55.5\scriptsize{ $\pm$ 0.6}      
                     &        
                     &--  &--         \\

                     & MTL~\cite{sun2019meta}                       
                     & {\color{blue}45.1\scriptsize{ $\pm$ 1.8}}       & {\color{blue}57.6\scriptsize{ $\pm$ 0.9}}      
                     &        
                     &--  &--         \\

                     & Transductive-ft~\cite{dhillon2019baseline}     
                     & 43.2\scriptsize{ $\pm$ 0.6}       & {\color{blue}57.6\scriptsize{ $\pm$ 0.6}}      
                     &        
                     &--  &--         \\


                   \midrule
                    & our baseline~(sec.~\ref{sec:baseline})         
                    & 40.84\scriptsize{ $\pm$ 0.71}              & 57.02\scriptsize{ $\pm$ 0.63}      
                    &       
                    & 71.71\scriptsize{ $\pm$ 0.86}              & 85.74\scriptsize{ $\pm$ 0.49}     
                    \\
                    
                    & adversarial       
                    & 43.44\scriptsize{ $\pm$ 0.71}                  & 58.69\scriptsize{ $\pm$ 0.56}        
                    &       
                    & 70.80\scriptsize{ $\pm$ 1.12}                  & 88.04\scriptsize{ $\pm$ 0.54}     
                    \\

                    & centroid           
                    & \textbf{45.83}\scriptsize{ $\pm$ 0.48}       & \textbf{59.74}\scriptsize{ $\pm$ 0.56}
                    &       
                    & \textbf{74.22}\scriptsize{ $\pm$ 1.09}        & \textbf{88.65}\scriptsize{ $\pm$ 0.55}     
                    \\
                    
                    \bottomrule
    \end{tabular}
    \\
    $^{\ddag}$ implementation from \cite{chen2019closer} for CUB, and from \cite{lee2019meta} for FC100 \quad 
    \label{tab:FC_CUB}
\end{table*}

%
\subsection{Cross-domain evaluation} 
We also evaluate our alignment strategies in cross-domain image classification. Here, following \cite{chen2019closer}, the base categories are drawn from \textit{mini}-ImageNet, but the novel categories are from CUB. 
As shown in table~\ref{tab:cross-domain}, we gain 1.3\% and 5.4\% over the baseline in the 1- and 5-shot, respectively, with our proposed centroid alignment. Adversarial alignment falls below the baseline in 1-shot by -1.2\%, but gains 5.9\% in 5-shot. Overall, our centroid alignment method shows absolute accuracy improvements over the state of the art (i.e., cosmax~\cite{chen2019closer}) of 3.8\% and 6.0\% in 1- and 5- shot respectively. We also outperform Robust-20~\cite{dvornik2019diversity}, an ensemble of 20 networks, by 4.65\% for 5-shot on \textit{mini}-ImageNet to CUB cross-domain.
One could argue that the three bird categories (i.e., house finch, robin, and toucan) in \textit{mini}-ImageNet bias the cross-domain evaluation. Re-training the approach by excluding these classes resulted in a similar performance as shown in table~\ref{tab:cross-domain}. 
\begin{table}[!t]
\centering
\setlength{\tabcolsep}{5pt}
\caption {Cross-domain results from \textit{mini}-ImageNet to CUB in 1-shot, 5-shot, 10-shot scenarios using a ResNet-18 backbone.} 
    \begin{tabular}{clccc} 
        \toprule
        & Method     & 1-shot     & 5-shot        & 10-shot \\    
        \midrule
	    & ProtoNet$^\ddag$~\cite{wang2018low}               & --                           & 62.02\scriptsize{ $\pm$ 0.70}      & -- \\   
	    & MAML$^\ddag$~\cite{finn2018probabilistic}         & --                           & 51.34\scriptsize{ $\pm$ 0.72}      & --\\   
	    & RelationNet$^\ddag$~\cite{sung2018learning}       & --                           & 57.71\scriptsize{ $\pm$ 0.73}      & -- \\  
	    & Diverse 20~\cite{dvornik2019diversity} & --                           & {\color{blue}66.17\scriptsize{ $\pm$ 0.73}}      & --\\


        & cosmax$^{\dag}$~\cite{chen2019closer}  & {\color{blue}43.06\scriptsize{ $\pm$ 1.01}}            & 64.38\scriptsize{ $\pm$ 0.86}           & {\color{blue}67.56\scriptsize{$\pm$0.77}}      \\

	    \midrule
         & our baseline (sec.~\ref{sec:baseline})         & 45.60\scriptsize{ $\pm$ 0.94}            & 64.93\scriptsize{ $\pm$ 0.95}           & 68.95\scriptsize{$\pm$0.78}     \\
         
        & adversarial                   & 44.37\scriptsize{ $\pm$ 0.94}             & 70.80\scriptsize{ $\pm$ 0.83}                   & 79.63\scriptsize{ $\pm$0.71}   \\
	    & adversarial$^*$               & 44.65\scriptsize{ $\pm$ 0.88}             & 71.48\scriptsize{ $\pm$ 0.96}                   & 78.52\scriptsize{ $\pm$0.70}  \\
	    
	    & centroid                      & 46.85\scriptsize{ $\pm$ 0.75}             & 70.37\scriptsize{ $\pm$ 1.02}                   & \textbf{79.98}\scriptsize{ $\pm$0.80}   \\ 
	    & centroid$^*$                  & \textbf{47.25}\scriptsize{ $\pm$ 0.76}    & \textbf{72.37}\scriptsize{ $\pm$ 0.89}          & 79.46\scriptsize{ $\pm$0.72}   \\
	    \bottomrule
    \end{tabular} \\ 
    $^*$ without birds (house finch, robin, toucan) in base classes \\
    $^{\dag}$ our implementation, with early stopping, \ \
    $^\ddag$ implementation from \cite{chen2019closer} 
    \label{tab:cross-domain} 
\end{table}

\section{Discussion} 
\label{sec:conclusion}
This paper presents the idea of associative alignment for few-shot image classification, which allows for higher generalization performance by enabling the training of the entire network, still while avoiding overfitting. 
To do so, we design a procedure to detect related base categories for each novel class. Then, we proposed a centroid-based alignment strategy to keep the intra-class alignment while performing updates for the classification task. We also explored an adversarial alignment strategy as an alternative. Our experiments demonstrate that our approach, specifically the centroid-based alignment, outperforms previous works in almost all scenarios.
%
%
The current limitations of our work provide interesting future research directions.  
First, the alignment approach (sec.~\ref{sec:alignment}) might include irrelevant examples from the base categories, so using categorical semantic information could help filter out bad samples. An analysis showed that $\sim$12\% of the samples become out-of-distribution (OOD) using a centroid nearest neighbour criteria on \textit{mini}ImageNet in 5-way 1- and 5-shot using ResNet-18. Classification results were not affected significantly by discarding OOD examples at each iteration. 
Second, the multi-modality of certain base categories look inevitable and might degrade the generalization performance compared to the single-mode case assumed by our centroid alignment strategy. Investigating the use of a mixture family might therefore improve generalization performance. 
Finally, our algorithms compute the related base once and subsequently keep them fixed during an episode, not taking into account the changes applied to the latent space during the episodic training. Therefore, a more sophisticated dynamic sampling mechanism could be helpful in the finetuning stage. 

  
%
\section*{Acknowledgement}
This project was supported by funding from NSERC-Canada, Mitacs, Prompt-Qu\'ebec, and E Machine Learning. We thank Ihsen Hedhli, Saed Moradi, Marc-Andr\'e Gardner, and Annette Schwerdtfeger for proofreading of the manuscript. 
\clearpage

\bibliographystyle{splncs04}
\bibliography{egbib}


\pagestyle{headings}
\mainmatter
\def\ECCVSubNumber{2826}  

\title{Associative Alignment \\ for Few-shot Image Classification \\ Supplementary Material} 

\titlerunning{Associative Alignment for Few-shot Image Classification Supp. Mat.}

%
\author{Arman Afrasiyabi$^*$, 
Jean-Fran\c{c}ois Lalonde$^*$, 
Christian Gagn\'e$^{* \dag}$ 
}
\authorrunning{A. Afrasiyabi et al.}
%
\institute{$^*$Universit\'e Laval, 
$^\dag$Canada CIFAR AI Chair, Mila   \\
\email{arman.afrasiyabi.1@ulaval.ca}\\
\email{\{jflalonde,christian.gagne\}@gel.ulaval.ca}  
\texttt{\url{https://lvsn.github.io/associative-alignment/}}}
\maketitle

\ \\
\ \\
\ \\
\ \\
In this supplementary material, the following items are provided:

\begin{enumerate}
    \item Validation error plot (sec.~\ref{sec:validation});  
    \item Ablation study on $B$ (sec.~\ref{sec:ablation-b}); 
    \item Visualization (sec.~\ref{sec:visualization}); 
    \item More ways (sec.~\ref{sec:more-ways});  
    \item Comparison to no alignment (sec.~\ref{sec:no-align});
    \item Sensitivity to wrongly-related classes (sec.~\ref{sec:sensitivity})
    \item Ablation on the margin (sec.~\ref{sec:ablation_on_m})
\end{enumerate}

 

\newpage

 \section{Validation error plot (refers to sec. 5.2)}
 \label{sec:validation}
 
\begin{figure}
\centering
\includegraphics[width=\linewidth]{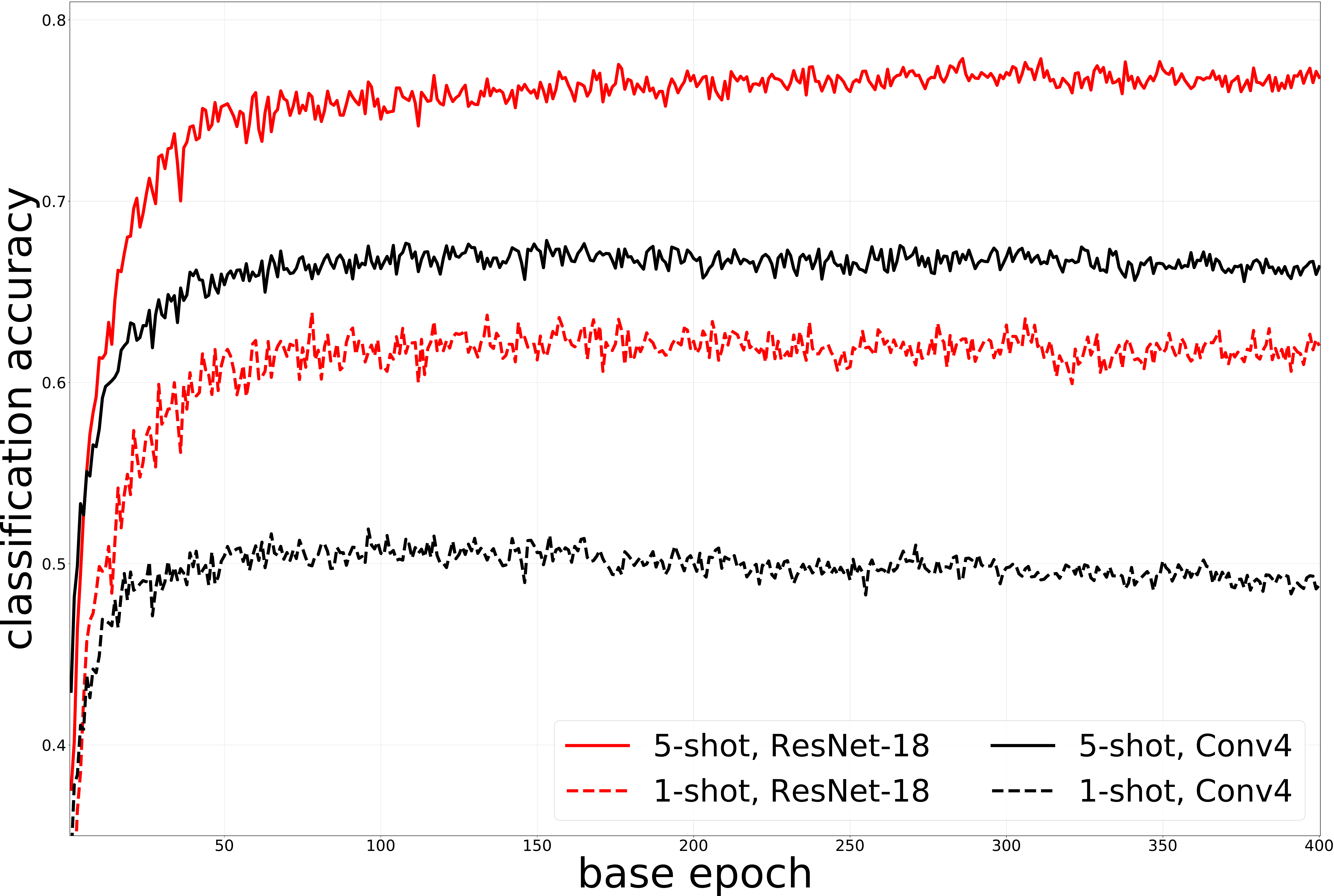}
\caption{Validation error after fine-tuning as a function of the number of pre-training base epochs on \textit{mini}-ImageNet with the cosmax loss. Pre-training for a fixed number of iterations (here 400 as in \cite{chen2019closer}) may lead to overfitting the feature extraction on the base set. Each curve represents the average of 50 episodes. 
}
\label{fig:overfitting}
\end{figure}

 Fig.~\ref{fig:overfitting} plots validation error after fine-tuning vs. the number of pre-training epochs. The ``cosmax'' function is used, with the entire network pre-trained on $\mathcal{X}^b$, and only the classification weights $\mathbf{W}$ fine-tuned on $\mathcal{X}^n$, as in~\cite{chen2019closer}. The decrease in accuracy over the epochs (after 150 epoch for 1-shot) shows that pre-training should not be conducted for a fixed number of epochs. 
 
 
\newpage

\section{Ablation study on $B$ (refers to sec. 6.1)}
\label{sec:ablation-b}

Table~\ref{tab:different_B} presents an ablation study for $B$, the number of related base classes selected for each novel class. We perform the study on few-shot image classification on the \textit{mini}-ImageNet dataset using ResNet-18 backbone. Overall, better results are obtained with a larger value of $B$, except for the adversarial alignment method in the 5-shot scenario. 


\begin{table*}[hbt!]
\centering
\caption{Effect of three different number of related bases $B$ on few-shot classification results on \textit{mini}-ImageNet using ResNet-18 backbones. 
  $\pm$ denotes the $95\%$ confidence intervals over 600 episodes.}
  \begin{tabular}{cc}
    \begin{tabular}{ccc}         
        \toprule
     
        $B $& 1-shot  & 5-shot \\ 
        \midrule
        arcm.   & 58.07 $\pm$ 0.82                 & 76.62  $\pm$ 0.58 \\
        1 &  55.76 $\pm$ 1.20                         & \textbf{79.34} $\pm$ 0.69      \\
        5 & 58.20 $\pm$ 1.14                          & 78.65 $\pm$ 0.94      \\
        10 & \textbf{58.84} $\pm$ 0.77                          & 77.92 $\pm$ 0.82     \\
        12 & 58.79 $\pm$ 0.81                          & 77.56 $\pm$ 0.85      \\
        \bottomrule
    \end{tabular}
    & 
    \begin{tabular}{ccc} 
        \toprule
        
        $B$ & 1-shot  & 5-shot \\ 
        \midrule
        arcm.   & 58.07 $\pm$ 0.82               & 76.62  $\pm$ 0.58 \\
        1 & 58.04 $\pm$ 0.98                          & 77.54 $\pm$ 0.73         \\   
        5 & 58.97 $\pm$ 1.06                          & 79.14 $\pm$ 0.91       \\   
        10 & \textbf{59.88} $\pm$ 0.67                  &\textbf{80.23} $\pm$ 0.73 \\
        12 & 60.04 $\pm$ 0.77                          & 80.18 $\pm$ 0.79      \\
        
        \bottomrule
    \end{tabular}
    \\
    (a) Adversarial alignment & (b) Centroid alignment
    \end{tabular}
    
    \label{tab:different_B}
\end{table*}

\newpage
\section{Visualization of the alignment methods (refers to sec. 6.2)}
\label{sec:visualization}
\begin{figure}[!h]
\footnotesize
\centering
\begin{tabular}{ccc} 
    \centering
    \footnotesize
    &  Centroid  & Adversarial  \\
    \rotatebox{90}{\hspace{4em} $B=1$} & 
    \includegraphics[width=.35\linewidth]{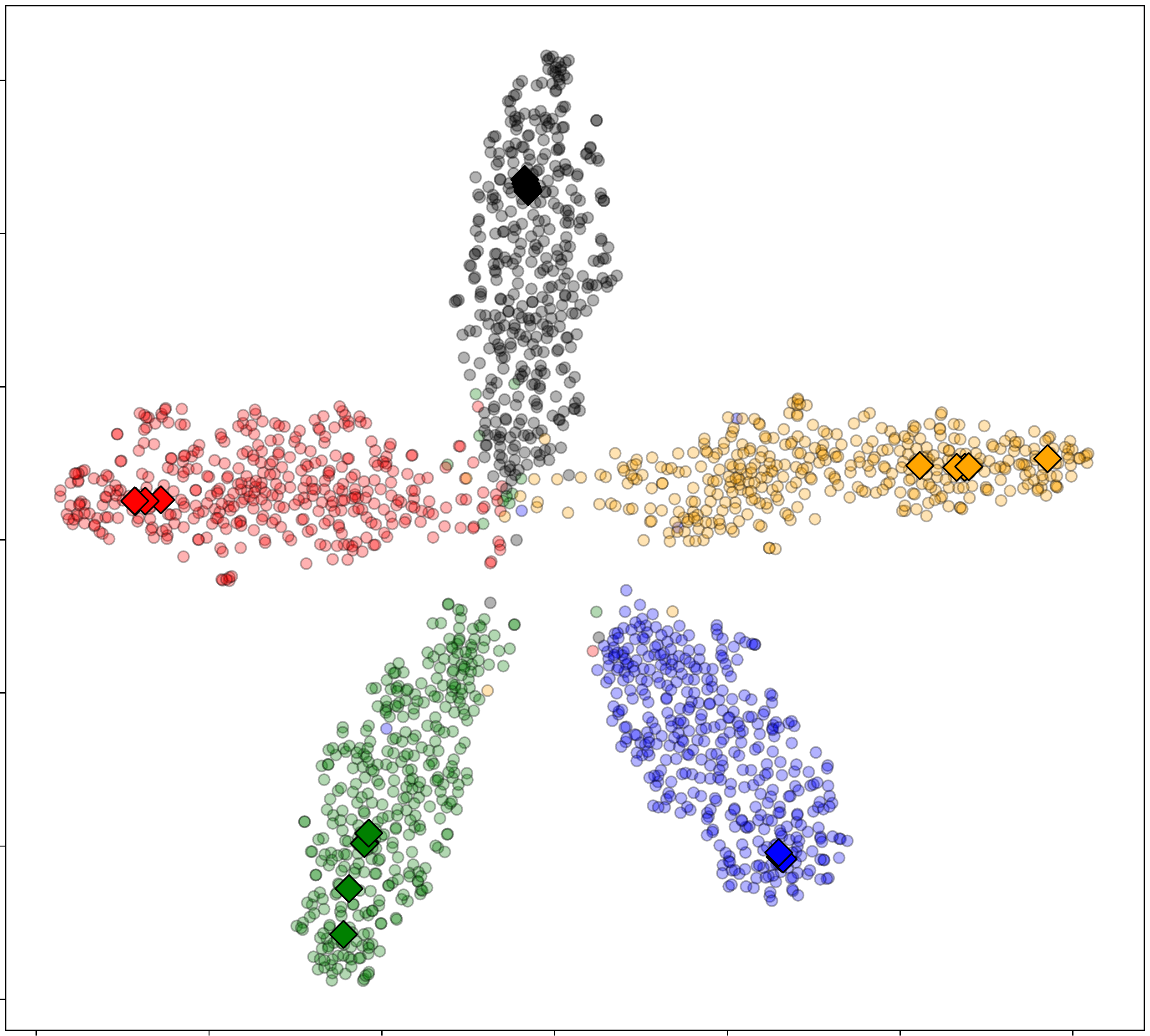} & 
    \includegraphics[width=.35\linewidth]{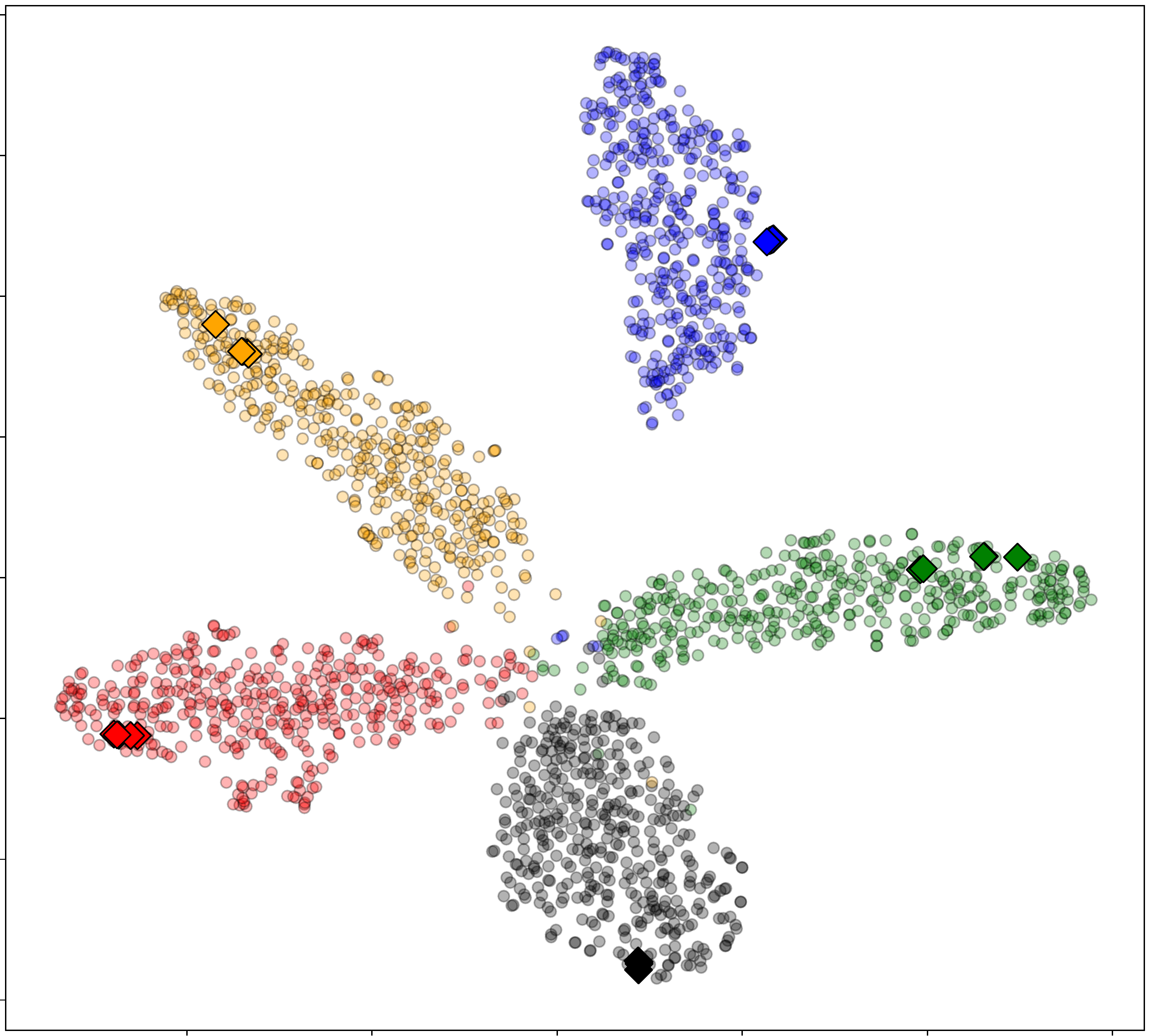} \\
    \rotatebox{90}{\hspace{4em} $B=10$} & 
    \includegraphics[width=.35\linewidth]{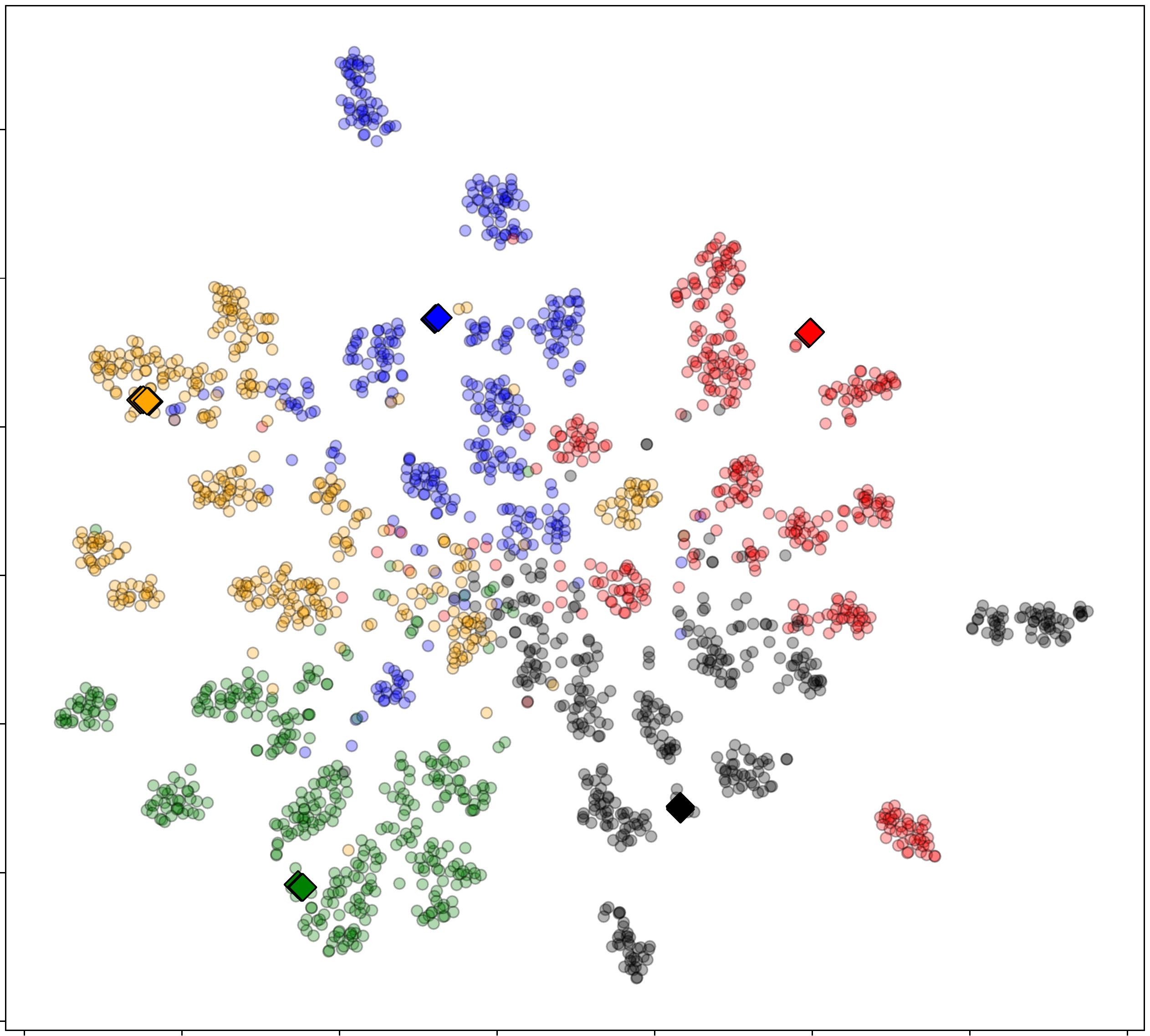} & 
    \includegraphics[width=.35\linewidth]{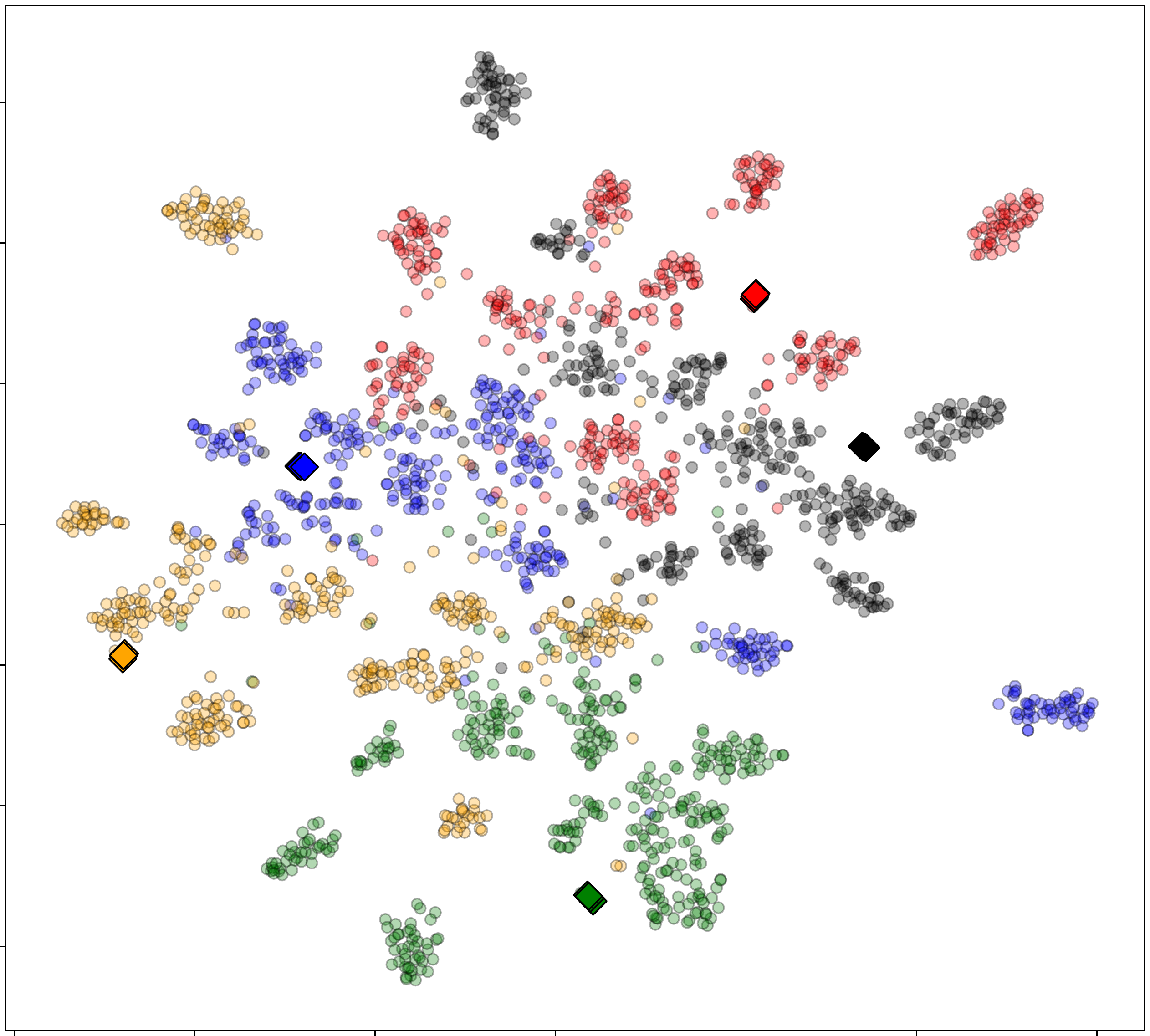} \\
    \multicolumn{3}{c}{
    \includegraphics[width=5cm,height=0.5cm]{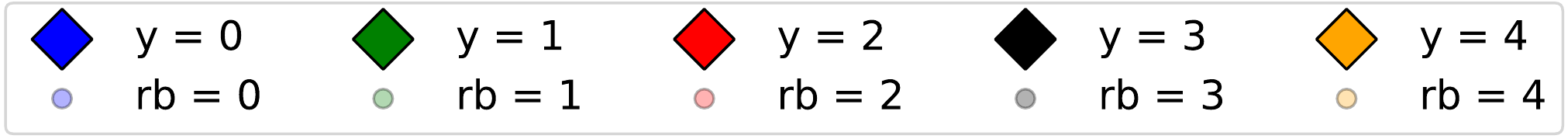}
    }
    \end{tabular}
    \caption{\label{fig:z_space_diff_B} \footnotesize{Aligning novel and related base classes. Columns present centroid and adversarial distribution matching while the rows compare picking $B=1$ and $B=10$ related base classes for each novel class. We use t-SNE~\cite{maaten2008visualizing} to visualize the 512-dimensional feature space of ResNet-18 in 2D. Results are for 5-shot in a 5-way setting.}}
\end{figure}

Fig.~\ref{fig:z_space_diff_B} presents a 2D visualization of our adversarial and centroid alignment methods using t-SNE~\cite{maaten2008visualizing} on \textit{mini}ImageNet (see sec. 6.1 for the dataset description) dataset in 5-shot 5-way scenario. While both methods achieve similar results with $B=1$, the centroid method results yields more discriminative class separation compared to the adversarial method with $B=10$. 
The multi-modalities of certain base categories look inevitable and might indeed degrade the generalization performance compared to the single-mode case assumed by our centroid alignment strategy. We compute the percentage of classes for which our centroid alignment approach: 1) improves, 2) does not change, or 3) deteriorates performance compared to our strong baseline (using a fixed threshold of 1\% on classification accuracy). In the 5-shot scenario using ResNet-18 on \textit{mini}-ImageNet, our centroid alignment approach results in improvements for 69.8\% of the classes (with 13.9\% not changing, and 16.3\% deteriorates).

\newpage
\section{More-way (refers to sec. 6.2)}
\label{sec:more-ways}
We experiment with N-way, 5-shot experiment (for N = 5, 10, 20) to examine the effect of associative alignment on more-way using \textit{mini}-ImageNet. As Table~\ref{tab:N-way} presents, our associative alignment gains on the compared meta-learning and standard transfer learning methods. Specifically, we outperform the best of the compared method by 6.67\%, 4.47\%, 3.82\% in 5-, 10-, and 20-way respectively. Note that we used 10, 5, 3 number of related base classes (B) 5-way, 10-way and 20-way respectively which corresponds to 60 classes out of all 64 base categories in \textit{mini}-ImageNet.

\begin{table*}[hbt!]
\centering
\caption{N-way 5-shot classification results on \textit{mini}-ImageNet using ResNet-18 backbone.   $\pm$ denotes the $95\%$ confidence intervals over 600 episodes.
The best results prior this work is highlighted in blue, and the best results are presented in boldfaced.} 
    \begin{tabular}{ rrccccccc}  
        \toprule
        
          & Method   &  & \textbf{5-way}  &  & \textbf{10-way}  &   & \textbf{20-way}            \\ 
        \midrule
        
    \multirow{3}{*}{\rotatebox{90}{meta-l.}}
         
        & MatchingNet$^\ddag$~\cite{vinyals2016matching}  &
        & 68.88 $\pm$ 0.69     &      & 52.27 $\pm$ 0.46    &        & 36.78 $\pm$ 0.25      
        \\

        & ProtoNet$^\ddag$~\cite{snell2017prototypical}    &
        & 73.68 $\pm$ 0.65         &         & 59.22 $\pm$ 0.44         &       &  44.96  $\pm$ 0.26           
        \\

        & RelationNet$^\ddag$~\cite{sung2018learning}       &  
        & 69.83 $\pm$ 0.68                                  &           
        & 53.88  $\pm$ 0.48                                 &        
        & 39.17 $\pm$ 0.25           
        \\      
 
        \midrule    
    
    \multirow{4}{*}{\rotatebox{90}{transfer-l.}}       
    
        & softmax~\cite{chen2019closer}     &
        & 74.27 $\pm$ 0.63                  &           
        & 55.00 $\pm$ 0.46                  &
        & 42.03 $\pm$ 0.25                  
        \\

        & cosmax~\cite{chen2019closer}     &
        & {\color{blue} 75.68 $\pm$ 0.63}                 &        
        & {\color{blue} 63.40 $\pm$ 0.44}                 &
        & {\color{blue} 50.85 $\pm$ 0.25}                 
        \\

        & our baseline (sec. 5.1)                    &
        & 76.62 $\pm$ 0.58                &
        & 62.95 $\pm$ 0.83                &
        & 51.92 $\pm$ 1.02                    
        \\
        
        \midrule
        & B{\ \ \ \ \ \ \ }  &
        & 10  &
        & 5   &
        & 3   &
        \ \\
        \midrule
     
    \multirow{2}{*}{\rotatebox{90}{align.}}

        & adversarial                          &
        &  77.92 $\pm$ 0.82                    &
        & 64.87 $\pm$ 0.96                     &
        & 52.46 $\pm$ 0.99                        
        \\

        & centroid       & 
        & \textbf{80.35} $\pm$ 0.73                    &        
        & \textbf{68.17} $\pm$ 0.79                             &
        & \textbf{54.67} $\pm$ 1.02                             
        \\
        \bottomrule
    \end{tabular}
    \\  
    $^\ddag$ implementation from \cite{chen2019closer}  
    \label{tab:N-way}
\end{table*}

\newpage
\section{Comparison to no alignment}
\label{sec:no-align}

Table~\ref{tab:big_picture} illustrates the effect of training the network using both novel and their related classes, but without the alignment losses. The results are shown in the ``no alignment'' row in table~\ref{tab:big_picture} below. Excluding the alignment loss slightly improves the accuracy compared to baseline by 0.82\% and 0.24\% in 1-shot and 5-shot using Conv4, respectively; however, it falls below the baseline by -2.13\% and -2.34\% in 1-shot and 5-shot using ResNet-18, respectively. In addition, except for the adversarial alignment in 1-shot using Conv4, both of the alignment strategies result in accuracy improvement in all of the scenarios, which shows the necessity of an alignment strategy. 

\begin{table*}[hbt!]
\centering
\caption{Evaluating the necessity of alignment loss. Few-shot classification results on \textit{mini}-ImageNet using both Conv4 and ResNet-18 backbones.  $\pm$ denotes the $95\%$ confidence intervals over 600 episodes.} 
    \begin{tabular}{ rrccccc} 
        \toprule
        & & \multicolumn{2}{c}{Conv4} & & \multicolumn{2}{c}{ResNet-18}   \\  
          &                           & 1-shot  & 5-shot  &  & 1-shot  & 5-shot \\ 
        \midrule
        \multicolumn{2}{r}{baseline} 
        & 51.90 $\pm$ 0.79      & 69.07 $\pm$ 0.62             &      & 58.07 $\pm$ 0.82            & 76.62 $\pm$ 0.58 \\
        \midrule
        
        \multicolumn{2}{r}{no alignment} 
        & 52.72 $\pm$ 0.79                  & 69.31 $\pm$ 0.69                &       & 55.94 $\pm$ 0.88                  & 74.28 $\pm$ 0.83     \\           
        \midrule
        
        \multirow{2}{*}{ \ \ \  alignment} 
        & adversarial     & 52.13 $\pm$ 0.99                  & 70.78 $\pm$ 0.60                &       & 58.84 $\pm$ 0.77                          & 77.92 $\pm$ 0.82     \\
        & centroid        &\textbf{53.14} $\pm$ 1.06          & \textbf{71.45} $\pm$ 0.72       &       & \textbf{59.88} $\pm$ 0.67                 & \textbf{80.35} $\pm$ 0.73 \\
        \bottomrule
    \end{tabular}
    \\
    \label{tab:big_picture}
\end{table*}



\newpage
\section{Sensitivity to wrongly-related classes}
\label{sec:sensitivity}
We also evaluate the sensitivity of the algorithm to the percentage of wrongly-related classes by replacing an increasing number of related base classes (selected by our algorithm) with random base classes instead (while keeping the total number of related base classes fixed to B=10). Results with the centroid alignment on \textit{mini}-ImageNet and ResNet-18 are shown in table~\ref{tab:sensitivity}. 

Small changes to the selected classes have little impact on performance showing the stability of our approach.
Replacing 5 randomly-selected base classes with random ones still results in improved performance in the 5-shot
scenario. Even if heuristic, our related base class selection algorithm results in much improved performance
compared to the 0/10 case. 

\begin{table*}[hbt!]
\centering
\caption{Evaluating the sensitivity to wrongly-related classes. Few-shot classification results on \textit{mini}-ImageNet using ResNet-18 backbone.  $\pm$ denotes the $95\%$ confidence intervals over 600 episodes.} 
    \begin{tabular}{ rrccccc}   
        \midrule
        & selected / random                 &          & 1-shot   &    & 5-shot  \\ 
        \midrule
        \multicolumn{2}{r}{[paper] 10 / 0}  & & \textbf{59.98} $\pm$ 0.7   &   & \textbf{80.35} $\pm$ 0.7           \\ 
        \multicolumn{2}{r}{9 / 1 }          & & 59.74 $\pm$ 0.7   &   & 80.07 $\pm$ 0.9           \\ 
        \multicolumn{2}{r}{8 / 2 }          & & 59.77 $\pm$ 0.6   &   & 78.69 $\pm$ 0.8           \\ 
        \multicolumn{2}{r}{5 / 5}           & & 58.36 $\pm$ 0.7   &   & 77.35 $\pm$ 0.8           \\ 
        \multicolumn{2}{r}{0 / 10}          & & 56.72 $\pm$ 1.2   &   & 76.19 $\pm$ 0.8           \\ 
        \multicolumn{2}{r}{[paper] baseline }& & 58.07 $\pm$ 0.8   &   & 76.62 $\pm$ 0.6           \\ 
        \bottomrule
    \end{tabular}
    \\ 
    \label{tab:sensitivity}
\end{table*}

\newpage
\section{Ablation on the margin $m$}
\label{sec:ablation_on_m}

We used episodic cross-validation to find the margin ($m$). In our experiments, we found that $m$ needs to be adjusted according to the architectures rather than the datasets, which is likely due to its relation to the network learning capacity. An ablation for $m$ on the \emph{mini}-ImageNet validation set for the 5-way scenario is presented in table~\ref{tab:margin_ablation}.

\begin{table*}[hbt!]
\centering
\caption{ablation for margin ($m$) on the \textit{mini}-ImageNet using ResNet-18 and Conv4 backbones.  $\pm$ denotes the $95\%$ confidence intervals over 600 episodes.} 
    \begin{tabular}{ rrcccccc}   
       \toprule
        & & & \multicolumn{2}{c}{Conv4} & & \multicolumn{2}{c}{ResNet-18}   \\      
        &  & $m$                           & 1-shot  & 5-shot & & 1-shot  & 5-shot \\ 
        \midrule
        \multicolumn{2}{r}{} &0.9       & 48.6  & 66.9  &  & 58.1    & 77.0      \\  
        \multicolumn{2}{r}{} &0.1       & 52.3   & 68.9  &  & 58.3    & 76.6    \\  
        \multicolumn{2}{r}{} &0.01      & 52.0   & 67.5  &  & 60.0    & 77.6    \\  
        \bottomrule
    \end{tabular}
    \\ 
    \label{tab:margin_ablation}
\end{table*}

\newpage

%
%
\end{document}